\documentclass[conference]{IEEEtran}
\IEEEoverridecommandlockouts
\IEEEpubid{\makebox[\columnwidth]{XXX-X-XXXX-XXXX-X/XX/\$XX.00~
\copyright2021 IEEE \hfill} \hspace{\columnsep}\makebox[\columnwidth]{ }}
\usepackage[hyphens]{url}  
\usepackage{graphicx} 
\usepackage{caption} 
\usepackage[switch]{lineno}
\usepackage{amsmath}
\usepackage{amssymb}
\usepackage{mathrsfs}
\usepackage{subcaption}
\usepackage{multirow}
\usepackage{makecell}
\usepackage{cite}
\usepackage{xcolor}
\usepackage{amsfonts}

\title{Towards Feature-free TSP Solver Selection: A Deep Learning Approach}

\author{
    \IEEEauthorblockN{
    Kangfei Zhao\IEEEauthorrefmark{1}\textsuperscript{\textsection}, 
    Shengcai Liu\IEEEauthorrefmark{2}\textsuperscript{\textsection\textreferencemark},
    Jeffrey Xu Yu\IEEEauthorrefmark{1}, 
    Yu Rong\IEEEauthorrefmark{3}}
    \IEEEauthorblockA{\IEEEauthorrefmark{1}The Chinese University of Hong Kong,
    \{kfzhao, yu\}@se.cuhk.edu.hk} 
    \IEEEauthorblockA{\IEEEauthorrefmark{2}Southern University of Science and Technology,
     liusc3@sustech.edu.cn} 
    \IEEEauthorblockA{\IEEEauthorrefmark{3}Tencent AI Lab,
     yu.rong@hotmail.com} 
}
\begin{document}

\maketitle
\begingroup\renewcommand\thefootnote{\textsection}
\footnotetext{Equal contribution}
\endgroup
\begingroup\renewcommand\thefootnote{\textreferencemark}
\footnotetext{Corresponding author}
\endgroup

\begin{abstract}
It is widely recognized that for the traveling salesman problem (TSP), there exists no universal best solver for all problem instances.
This observation has greatly facilitated the research on Algorithm Selection (AS), which seeks to identify the solver best suited for each TSP instance.
Such segregation usually relies on a prior representation step, in which problem instances are first represented by carefully established problem features.
However, the creation of good features is non-trivial, typically requiring considerable domain knowledge and human effort.
To alleviate this issue, this paper proposes a deep learning framework, named CTAS, for TSP solver selection.
Specifically, CTAS exploits deep convolutional neural networks (CNN) to automatically extract informative features from TSP instances and utilizes data augmentation to handle the scarcity of labeled instances.
Extensive experiments are conducted on a challenging TSP benchmark with 6,000 instances, which is the largest benchmark ever considered in this area.
CTAS achieves over 2$\times$ speedup of the average running time, compared with the single best solver.
More importantly, CTAS is the first feature-free approach that notably outperforms classical AS models, showing huge potential of applying deep learning to AS tasks. 
\end{abstract}

\section{Introduction}

The Traveling Salesman Problem (TSP) is a well-known NP-hard combinatorial optimization problem.
Given $n$ cities, it aims to find a tour that traverses each city exactly once with the shortest traveling distance in total.
TSP has broad applications in many disciplines, from routing planning in urban computing to circuit layout in chip manufacturing, from DNA sequencing in bioinformatics to telescope scheduling in astronomy.
In this paper, we consider a commonly studied TSP, i.e., the Euclidean TSP, in which the cities correspond to points in the 2-dimensional Euclidean plane and the distances are Euclidean distances.

Over the past decades, a number of advanced TSP solvers have been proposed to find high-quality solutions.
However, like other computationally hard problems, for TSP it is well recognized that there exists no universal best solver dominating all other solvers on all possible instances \cite{DolpertM97,KerschkeKBHT18}.
In other words, different TSP solvers are good at solving different instances.
This observation has incited many researchers to exploit the complementarity between different TSP solvers, by including them into a so-called algorithm portfolio. 
It is then expected that an instance would be satisfactorily handled by at least one solver in the portfolio.

Generally, there exist two common strategies to utilize a given algorithm portfolio.
One is parallel running strategy \cite{liu2017experience,LiuT019,LiuT020gen,tang2020few}, which runs all solvers independently in parallel on a given instance until the first of them solves it.
The main drawback of this strategy is that it causes a certain waste of computational resources, since running only the best-performing solver on the instance could already achieve the same performance.
The other strategy, called algorithm selection (AS) \cite{rice1976algorithm}, could avoid such issue by only selecting the best suited solver for the instance to run.
Specifically, the solver selector is usually built by training data-driven machine learning models offline, with a feature set for each training instance collected in advance.

Despite the tremendous progress achieved in the development of AS approaches (not just for TSP solver selection, see \cite{Kotthoff14}),
such a feature-based paradigm typically suffers from several main drawbacks.
First, it fundamentally relies on a set of informative structural features, designed by human experts, to distinguish amongst the instances.
Over the years, many problem domains (including TSP) have accumulated their own set of features, yet at the core they are mostly a collection of everything that was considered useful in the past \cite{KerschkeKBHT18}.
This means a pain-striking feature engineering is necessary to filter redundant, irrelevant and conflict features from a large feature pool, to obtain really useful features for a particular selection technique and dataset.
Once the dataset or selection model changes, the whole procedure needs redoing again.
Second, the expressive power of the statistical models, which are commonly used by existing AS approaches, is somehow limited.
For fast-growing instance volumes and diversity, the statistical models face the risk of underfitting.
The above consideration motivates us to build an end-to-end deep learning framework for TSP solver selection.
By end-to-end deep learning, it learns the features together with the solver selectors by directly encoding TSP instances into a latent representation using a powerful optimizer (e.g., stochastic gradient descent), thus eliminating the human element from the feature generation process.

The main contributions of this paper are three-fold.
First, we investigate several deep learning models and propose a convolution-based TSP solver selection (CTAS) framework to leverage the complementarity of TSP solvers.  
In particular, CTAS utilizes the convolution operator to automatically extract features from a 2-dimensional plane, which is the density map of the points in a TSP instance.
Due to its great expressive power, CTAS can capture the spatial patterns related to the performance of TSP solvers from local to global spatial density layer by layer.
Furthermore, to handle the scarcity of labeled instances and facilitate better generalization, we present safe data augmentation strategies that are specific for the TSP solver selection task.
Second, to fully assess the potential of CTAS, we generate a huge TSP benchmark consisting of 6,000 instances following six different distributions, which is the largest benchmark ever considered in this area.
Third, we conduct extensive experiments to demonstrate the effectiveness of CTAS. 
Compared with the single best solver, CTAS is able to achieve over 2$\times$ speedup of the average running time.
More importantly, as a feature-free approach solely based on the visual representation of the problem instances, CTAS surpasses the state-of-the-art statistical models for the first time in the literature.
The TSP benchmark and our implementation is publicly available at \url{https://github.com/Kangfei/TSPSelector}.

\section{Related Work}  
\label{sec:rw}
Over the last two decades, AS has been successfully applied to many computationally hard problems.
Machine learning techniques are the mainstream approaches for AS, and two strategies, namely classification and regression, are used.
More specifically, classification models predict the solver for each instance, whereas the regression models first estimate the performance of each solver, and then select the one with the best estimated performance.
A comprehensive survey on AS can be found in \cite{Kotthoff14}.

As a sub-field of AS, TSP solver selection has been investigated by several prior studies.
For examples, different classifiers were employed in \cite{KandaCHS11} and \cite{PiheraM14} to predict the solver to run, considering different feature sets.
A solver ranking model was trained in \cite{KandaCHSB16} to predict the rankings among the solvers, with the highest-ranking solver being selected.
A recent study \cite{KerschkeKBHT18} utilized both classification and regression strategies to build TSP selectors, and made an in-depth investigation of feature selection on different feature sets.
The experiment results in \cite{KerschkeKBHT18} showed that the learned selectors significantly outperformed the single best solver, thus achieving an advance in the state-of-the-art in solving TSP.
To the best of our knowledge, there exists only one study \cite{SeilerPBKT20} that attempted to utilize deep learning to TSP solver selection.
In particular, the authors used CNNs to select the best TSP solver, based on different visual representations of TSP instances including point clouds, minimum spanning tree and the nearest neighbor graph.
However, the experiment results in \cite{SeilerPBKT20} showed that the deep learning-based approach still fell behind classical feature-based AS models.
In this paper, we develop a novel feature-free selection framework based on the grid representation of the TSP instances, and the experiments demonstrate that our approach consistently outperforms the previous approach.

On the other hand, in the general area of AS, there has been growing research interest in leveraging deep learning to avoid the tedious work of feature engineering.
For SAT solver selection, a CNN was utilized in \cite{LoreggiaMSS16} to extract a set of features from images generated from the text files of the problem instances.
The experiment results in \cite{LoreggiaMSS16} are appealing, although the proposed approach could not yet achieve the performance of classical models.
AS approaches equipped with deep models were also applied to problem domains like 1-D bin packing \cite{AlissaSH19}, autonomous agent search in video game \cite{SigurdsonB17} and domain-independent planning \cite{Sievers0SSF19}.
The experiment results in these studies showed that using deep learning could often outperform the single best solver in the portfolio.
However, no results of comparison with classical AS models are provided in these works.

\section{Problem Statement}
\label{sec:pr}
Given a TSP instance set $\mathcal{I} = \{I_1, I_2, \cdots I_n\}$, and a TSP solver set
$\mathcal{A} = \{A_1, A_2, \cdots A_m\}$,
the TSP solver selection problem is to find a per-instance mapping $s: \mathcal{I} \rightarrow \mathcal{A}$ that optimizes its overall performance on $I$, according to a given performance metric $p$.
In this paper, $p$ is the penalized average running time with penalty factor 10 (PAR10)\cite{KerschkeKBHT18}, a widely-used hybrid performance metric.
Specifically, PAR10 is the average running time on all the instances in $\mathcal{I}$, where those unsuccessful runs (unable to find a solution of accepted quality level within the cut-off time) are counted as ten times the cut-off time.

To evaluate the solver selector $s$, as usually done in the literature, it needs to compare with two baselines, namely, the virtual best solver (VBS) and the single best solver (SBS).
VBS is the perfect selector which always selects the best solver for each instance in $\mathcal{I}$ without any selecting cost.
It provides an upper bound for the performance of any algorithm selector; due to imperfect selection and the cost incurred by selecting, no solver selectors can achieve VBS in practice.
On the other hand, SBS is the single solver in $\mathcal{A}$ that has the minimum PAR10 over $\mathcal{I}$.
In other words, SBS implicitly represents the trivial solver selector which always selects the same solver for all instances.
Therefore, an algorithm selector is only useful if it performs better than SBS.

\section{Convolution-based TSP Solver Selection}
\captionsetup[sub]{font=small}
\begin{figure*}[t]
\centering
\begin{subfigure}[b]{0.3\columnwidth}
  \centering
  \scalebox{1.0}{\includegraphics[width=\linewidth]{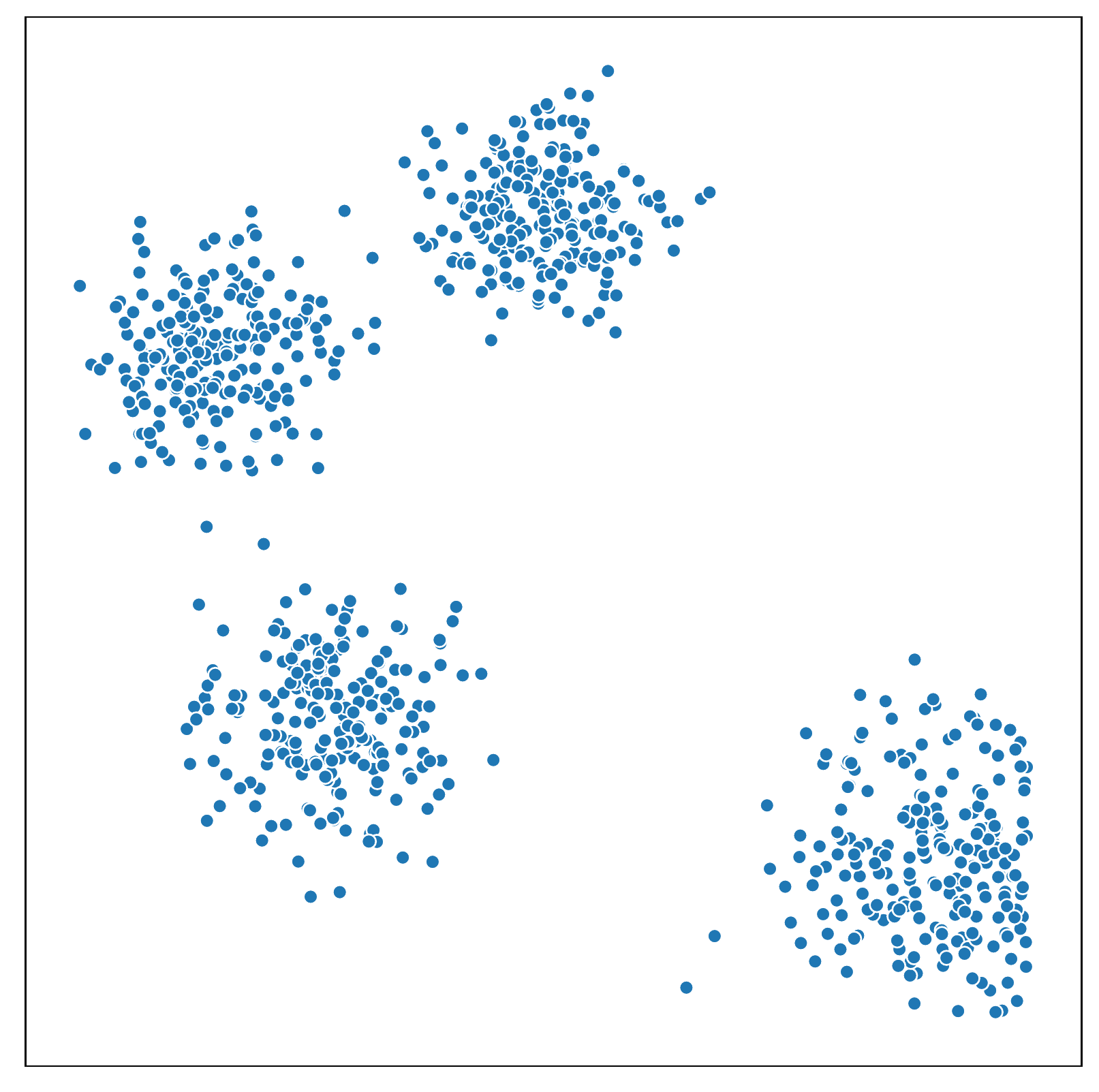}}
  \caption{2-D Points}
  \label{fig:trans:raw}
\end{subfigure}
\hfill
\begin{subfigure}[b]{0.3\columnwidth}
  \centering
  \scalebox{1.0}{\includegraphics[width=\linewidth]{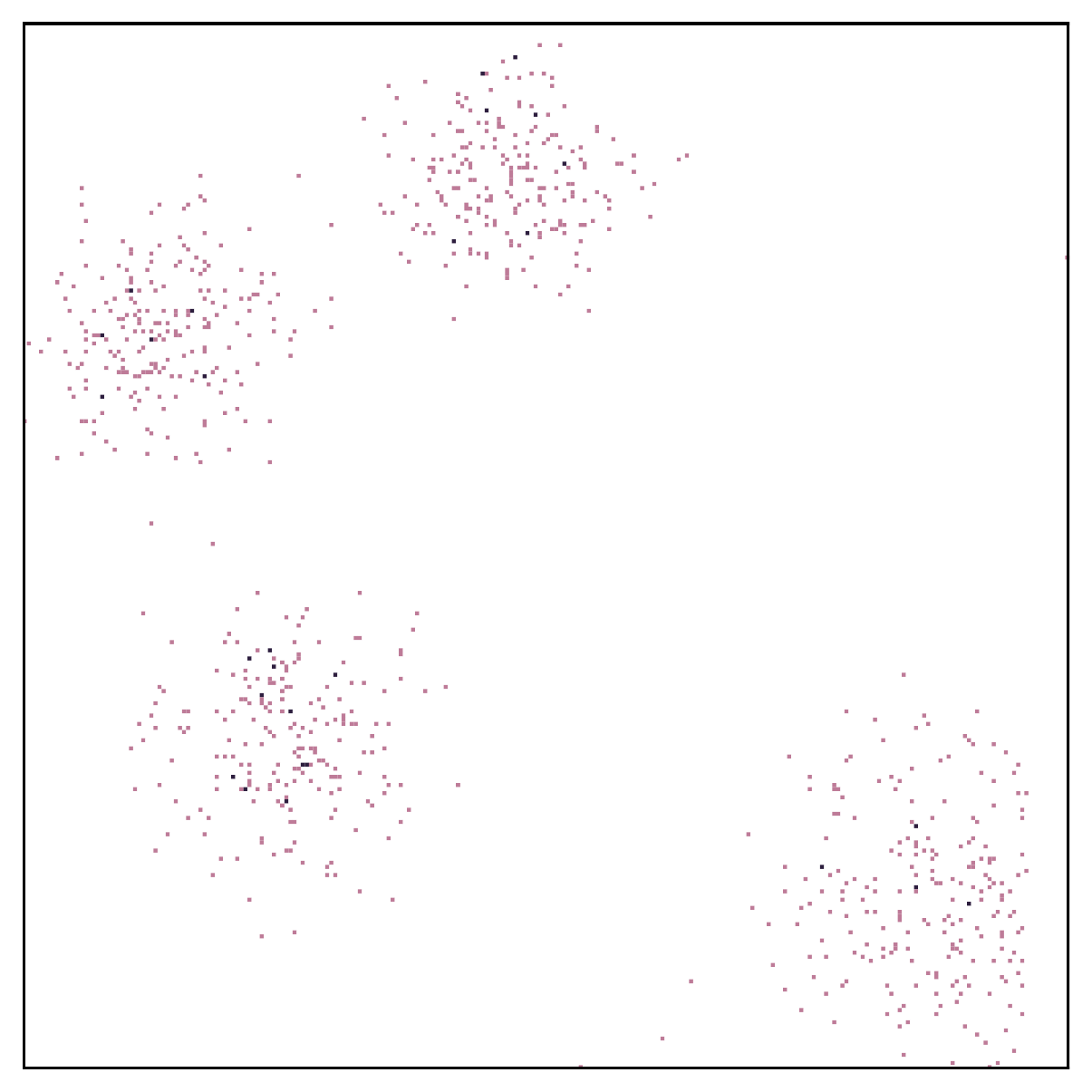}}
   \caption{Gridding}
   \label{fig:trans:grid}
\end{subfigure}
\hfill
\begin{subfigure}[b]{0.3\columnwidth}
  \centering
  \scalebox{1.0}{\includegraphics[width=\linewidth]{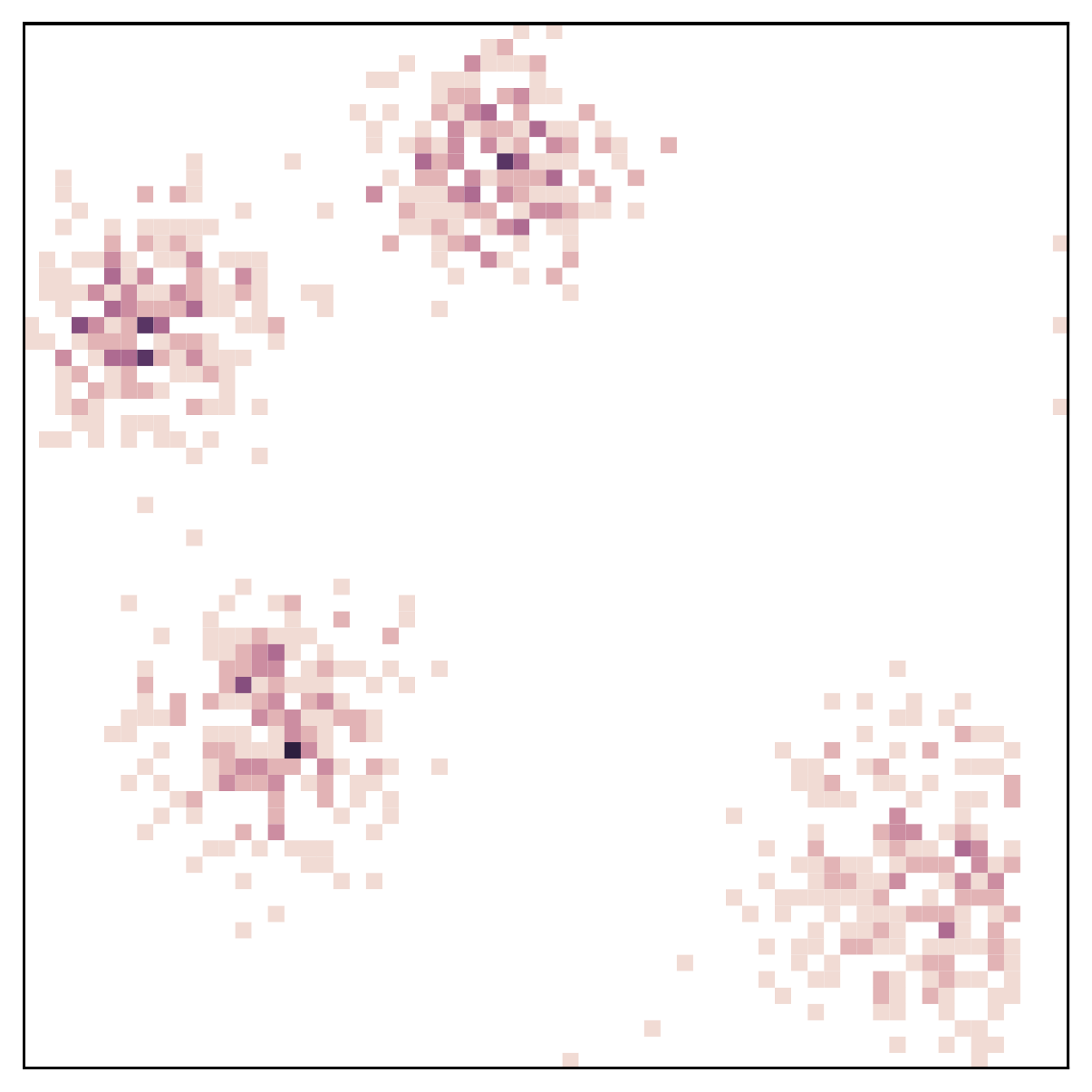}}
   \caption{Up-scaling}
   \label{fig:trans:interpolate}
\end{subfigure}
\hfill
\begin{subfigure}[b]{0.3\columnwidth}
  \centering
  \scalebox{1.0}{\includegraphics[width=\linewidth]{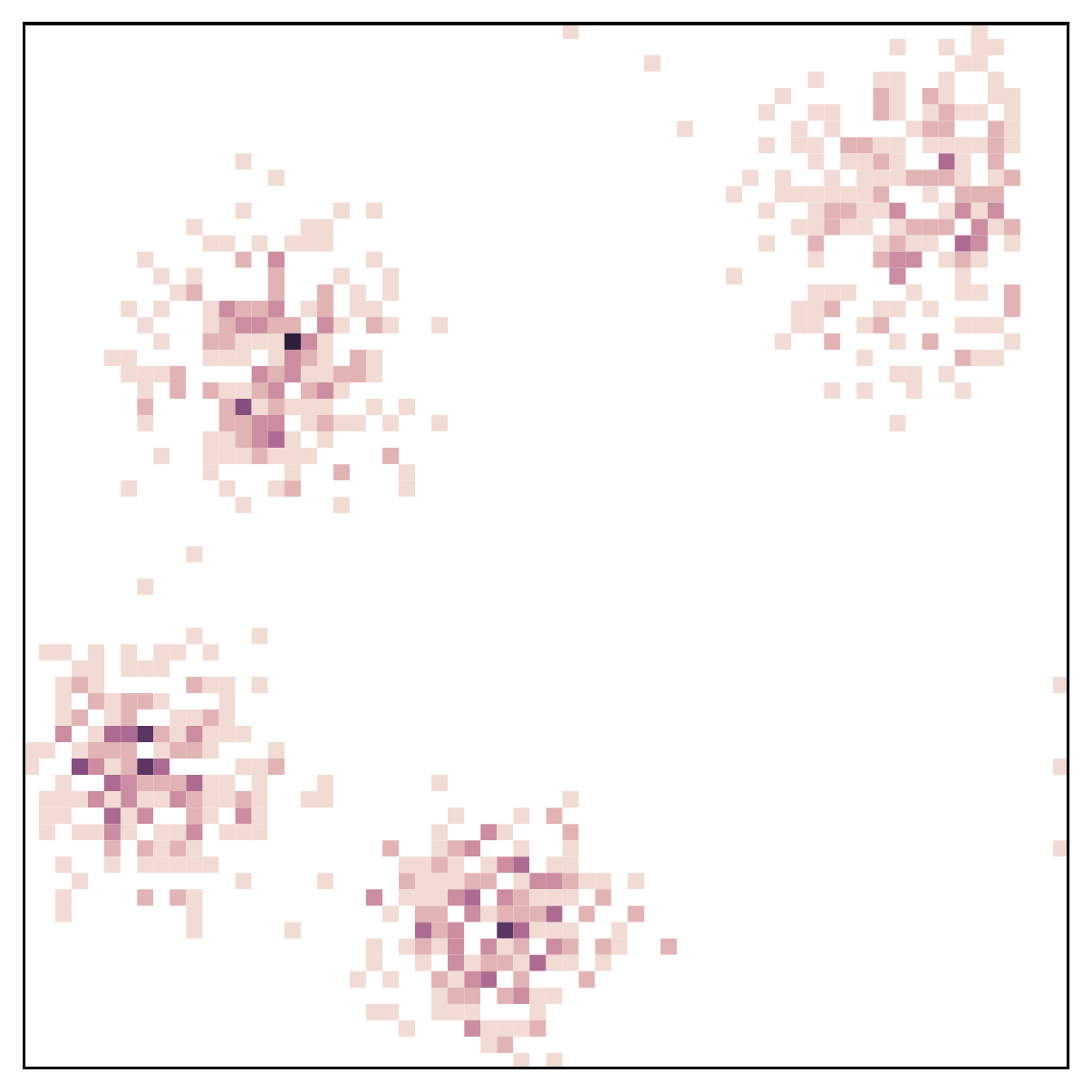}}
   \caption{Horizontal Flip}
   \label{fig:trans:hflip}
\end{subfigure}
\hfill
\begin{subfigure}[b]{0.3\columnwidth}
  \centering
  \scalebox{1.0}{\includegraphics[width=\linewidth]{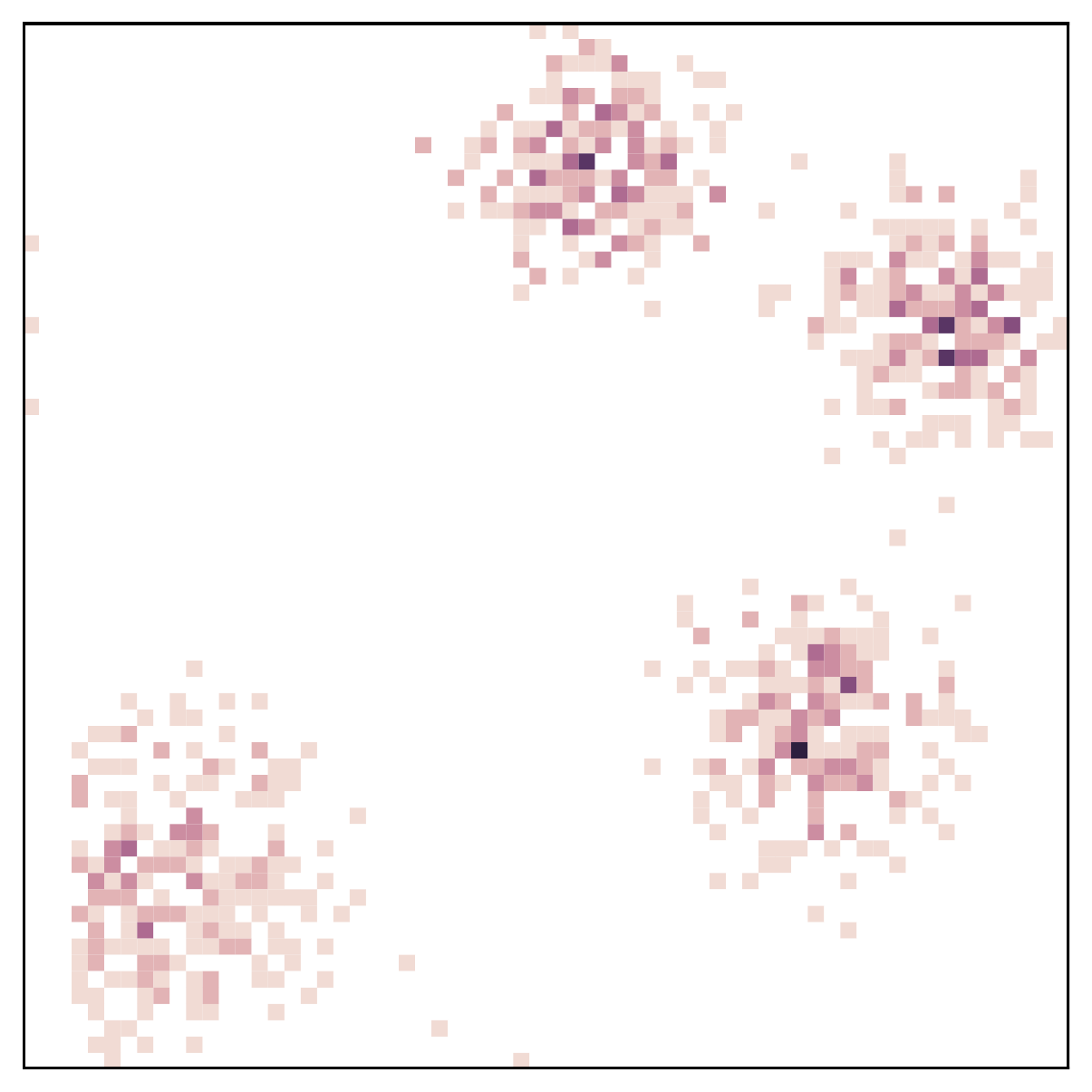}}
   \caption{Vertical Flip}
   \label{fig:trans:vflip}
\end{subfigure}
\hfill
\begin{subfigure}[b]{0.3\columnwidth}
  \centering
  \scalebox{1.0}{\includegraphics[width=\linewidth]{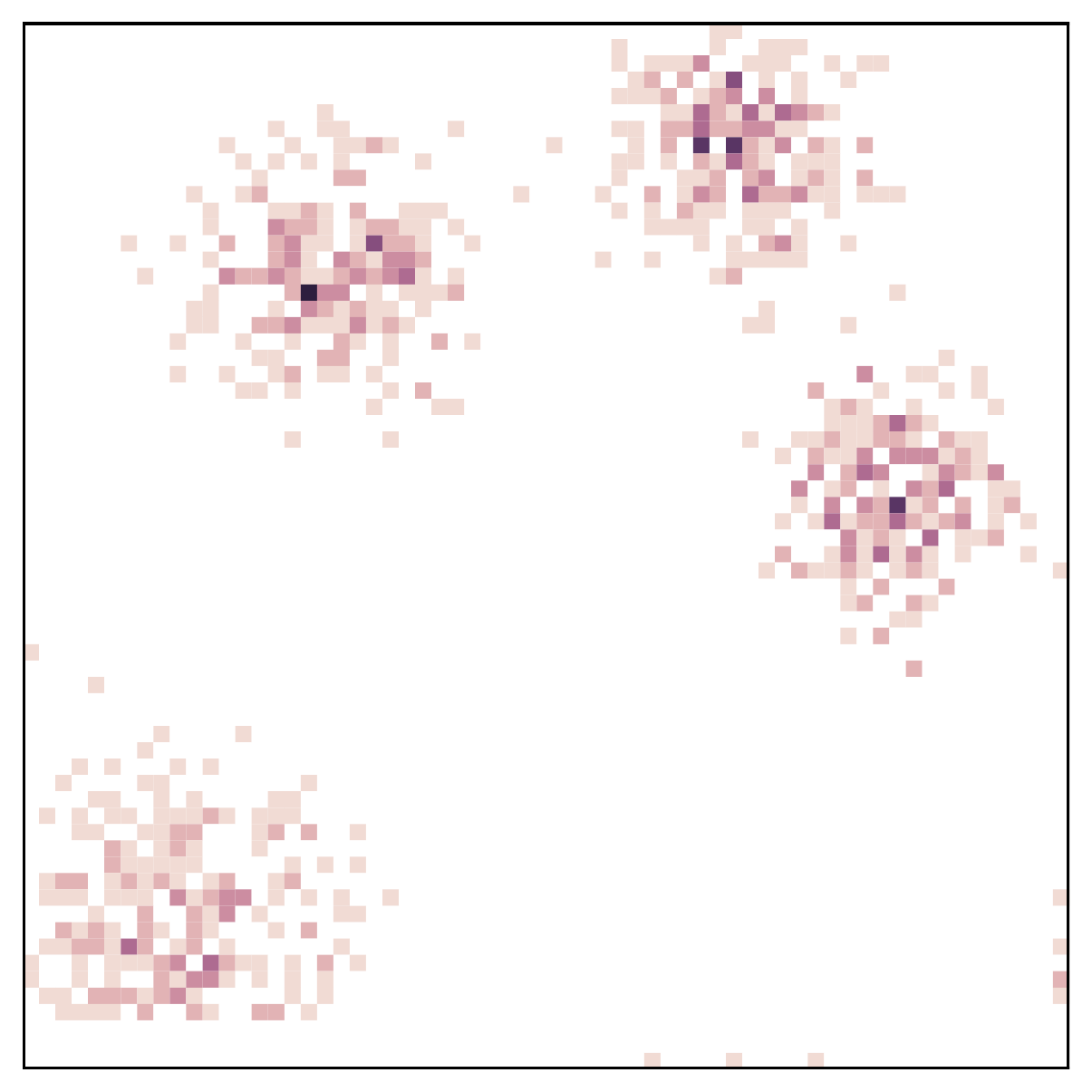}}
   \caption{Rotation}
   \label{fig:trans:rotate}
\end{subfigure}
\caption{An example of TSP data transformation.}
\label{fig:trans}
\end{figure*}

There are two prevailing models, the graph convolutional network (GCN)~\cite{KipfW17} and the convolutional neural network (CNN)~\cite{LeCunKF10}, that can be adopted to learn a TSP solver selector, as the models are designed to handle geometric and spatial data (e.g., TSP data).
We explored both options, and finally adopted CNN.
The reason why GCN might not perform well is explained below.
First, the graph convolution layer of GCN is regarded as a special Laplacian smoothing on the node features for new feature generation.
However, the points in a TSP instance do not have hands-on features except their coordinates. 
Second, when the input is a complete graph (e.g., a TSP instance), GCN degenerates into a graph network variant \textit{Deep Sets}~\cite{ZaheerKRPSS17}, which would severely suffer from the over-smoothing issue \cite{LiHW18} as the model aggregates the feature of all points.
Third, as the time complexity of training a GCN is linear to the number of edges, considering many TSP instances of interest are with hundreds even thousands of points, it is almost impossible to train a GCN even through some sampling-based models.
Below we elaborate on each component of the convolution-based approach (CTAS).

\subsection{From TSP Instance to Image}
Given a TSP instance of $n$ points $\{v_1, v_2, \cdots, v_n\}$, where each point $v_i$ has a coordinate in the 2-dimensional Euclidean plane.
The raw coordinates are first re-scaled by min-max normalization, after which all points would fall into a  $(0, 1)^2$ square, as shown in  Fig.~\ref{fig:trans:raw}.
Then the instance is transformed to a density map by gridding, with the spatial connection among points preserved.
Specifically, by uniformly and evenly dividing the $(0, 1)^2$ square into $c \times c$ cells, the density map is constructed as a $c \times c$ image, where the greyscale of pixel $(i, j)$ is the number of points falling into $\mathrm{cell}(i, j)$.
Fig.~\ref{fig:trans:grid} shows this discretization with $c=256$.
A potential problem of the density map is that it tends to be sparse (e.g., Fig.~\ref{fig:trans:grid}), where most cells are empty.
To alleviate this issue, we reduce $c$ to 64 and meanwhile up-scales the image with neighborhood interpolation.
Such enhancement also improves the resolution of the image, which is favorable for CNN to learn features via deep layers.
Fig.~\ref{fig:trans:interpolate} shows up-scaling a $64 \times 64$ density map by 4 times, resulting in a $256 \times 256$ image.
Compared to Fig.~\ref{fig:trans:grid} of the same resolution, the up-scaled image is clearer and sharper.

\subsection{Loss Function}
\label{se:losschoices}
Given $n$ TSP instances $\mathcal{I} = \{I_1, I_2, \cdots I_n\}$ and $m$ solvers $\mathcal{A} = \{A_1, A_2, \cdots A_m\}$, let $\mathcal{T} = \{\bf{t_1},\bf{t_2}, \cdots, \bf{t_n}$\}, where $\bf{t_i} \in \mathbb{R+}^{m}$ is an $m$-element vector such that $t_{ij}$ is the running time of solving instance $I_i$ by solver $A_j$.
We aim at building a model which maps a TSP instance represented by a density map to its best solver.
One straightforward strategy is to treat $m$ solvers as $m$ classes with which we train a CNN classifier.
The typical cross-entropy loss (on a single instance $I_i$) of the ``classification'' strategy is:
\begin{linenomath}
\begin{equation}
\label{eq:loss:ce}
L_{CE}(I_i) = \sum_{j = 1}^{m} w_{ij} \cdot p_{ij} \cdot \log(q_{ij}).
\end{equation}
\end{linenomath}
${\bf p_{i}} \in [0, 1]^m$ is the empirical distribution of selecting the $m$ solvers, which is a function of ${\bf t_{i}}$;
${\bf q_{i}} \in [0, 1]^m$ is the predicted likelihood, which is the output of CNN followed by the softmax function;
${\bf w_i} \in \mathbb{R}^m$ is an optional weight vector.
However, there exist several potential issues of applying Eq.~(\ref{eq:loss:ce}).
First, there are TSP instances on which multiple solvers perform similarly best.
For these instances, setting ${\bf p_{i}}$ to 0-1 hard label for one of them will neglect the other best candidate solvers, while setting ${\bf p_{i}}$ to probabilistic soft label will blur the supervision of the label.
Second, it does not explicitly meet the objective of the TSP solver selection task, i.e., minimizing the PAR10.
Therefore we also consider the ``regression'' strategy that directly leverages the differences of the running time in the learning process.
Specifically, a CNN regressor is trained to predict the running time of the $m$ solvers, by minimizing the mean squared error (MSE):
\begin{linenomath}
\begin{equation}
\label{eq:loss:mse}
L_{MSE}(I_i) = \sum_{j = 1}^{m} w_{ij} \cdot \left( q_{ij} - t_{ij} \right)^2.
\end{equation}
\end{linenomath}
Then the solver with the best estimated performance will be selected.
In Eq.~(\ref{eq:loss:mse}), the output of CNN, ${\bf q_i} \in \mathbb{R}^m$ is the estimated running time of the $m$ solvers on the instance $I_i$.
Optionally, ${\bf w_i} \in \mathbb{R}^m$ is the weight vector regarding $I_i$.

\captionsetup[sub]{font=small}
\begin{figure*}[t]
\centering
\begin{subfigure}[b]{0.3\columnwidth}
   \includegraphics[width=\linewidth]{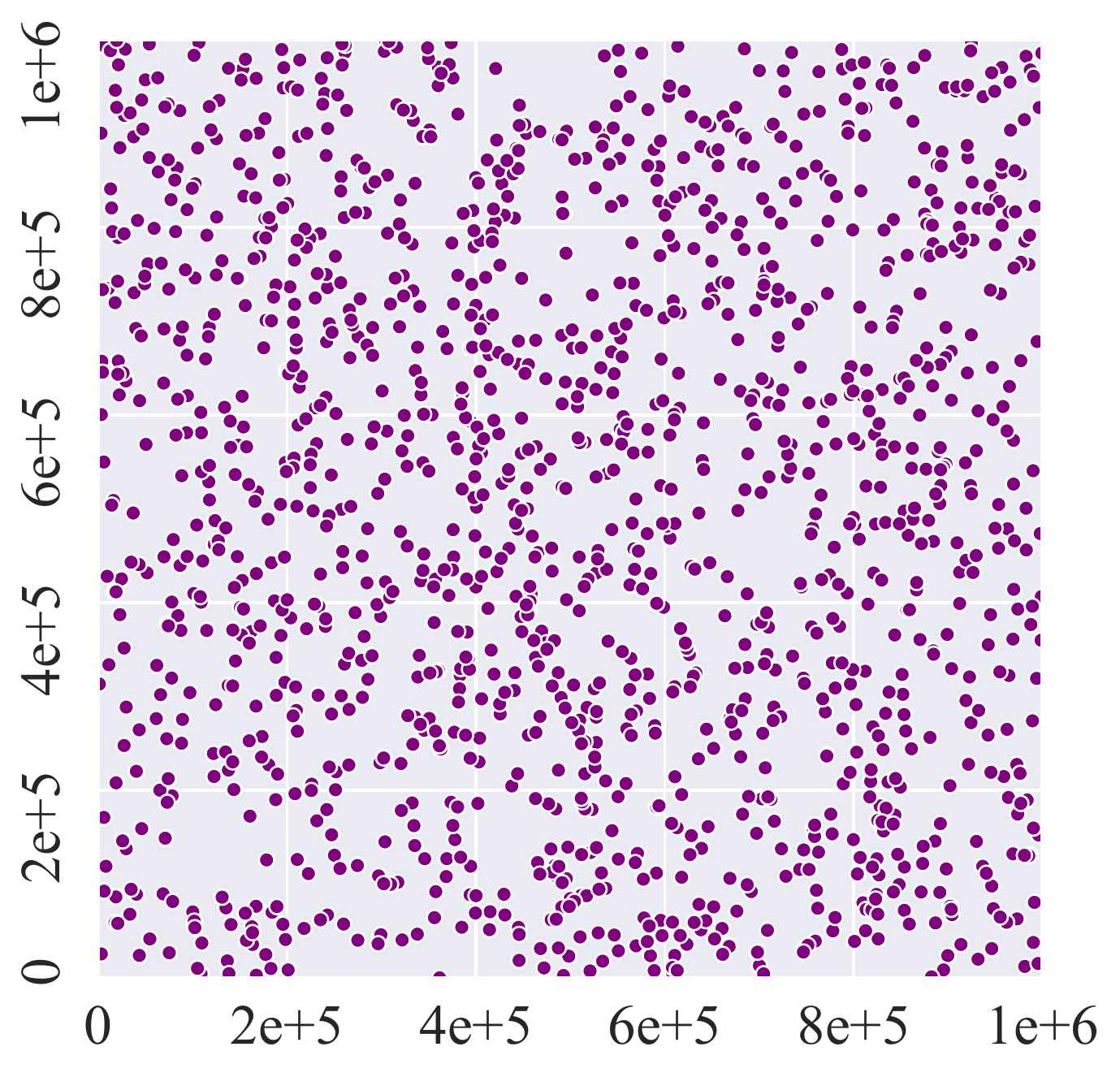}
   \caption{rue}
   \label{fig:tsp:rue}
\end{subfigure}
\hfill
\begin{subfigure}[b]{0.3\columnwidth}
   \includegraphics[width=\linewidth]{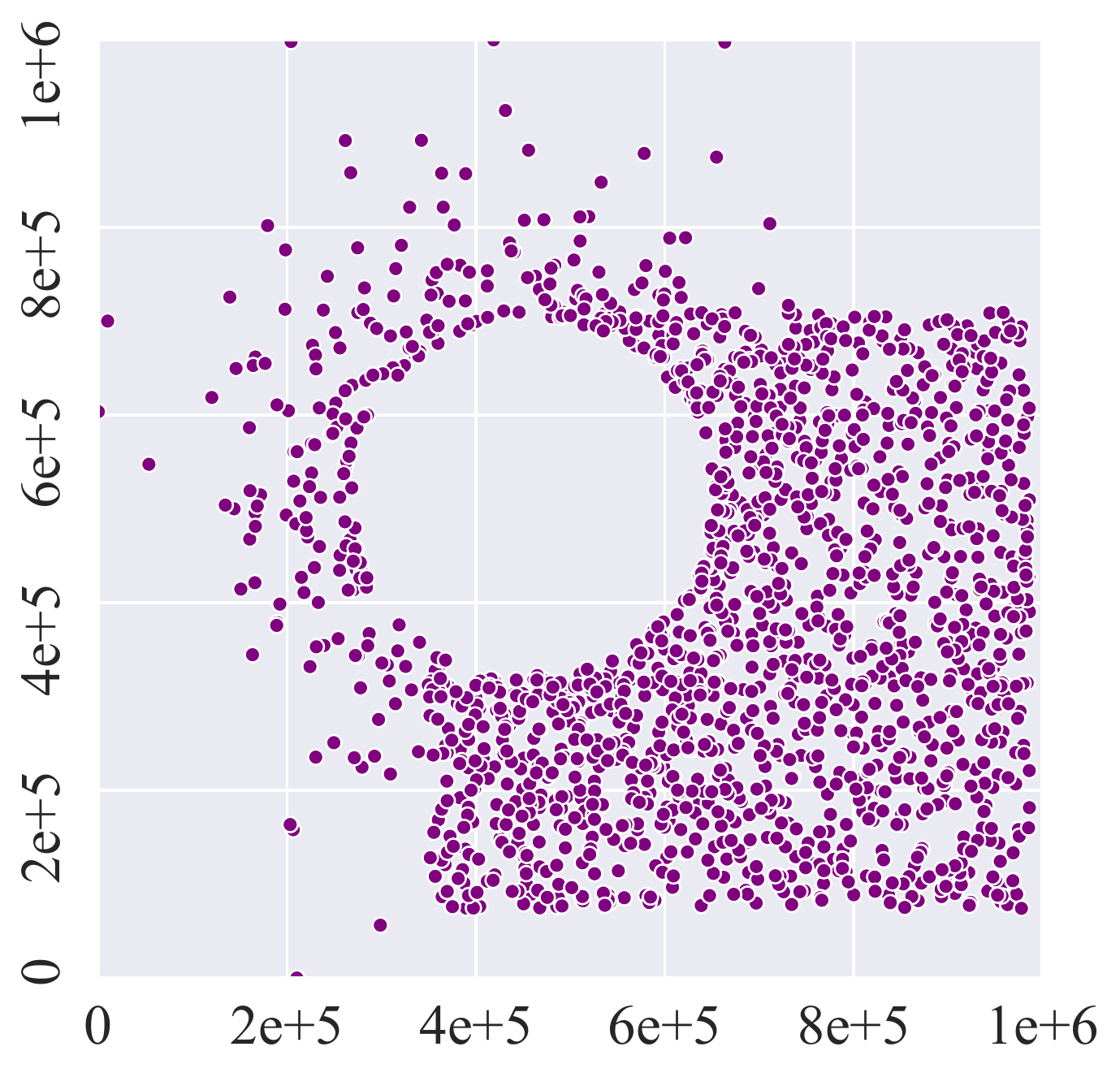}
   \caption{explosion}
   \label{fig:tsp:explosion}
\end{subfigure}
\hfill
\begin{subfigure}[b]{0.3\columnwidth}
   \includegraphics[width=\linewidth]{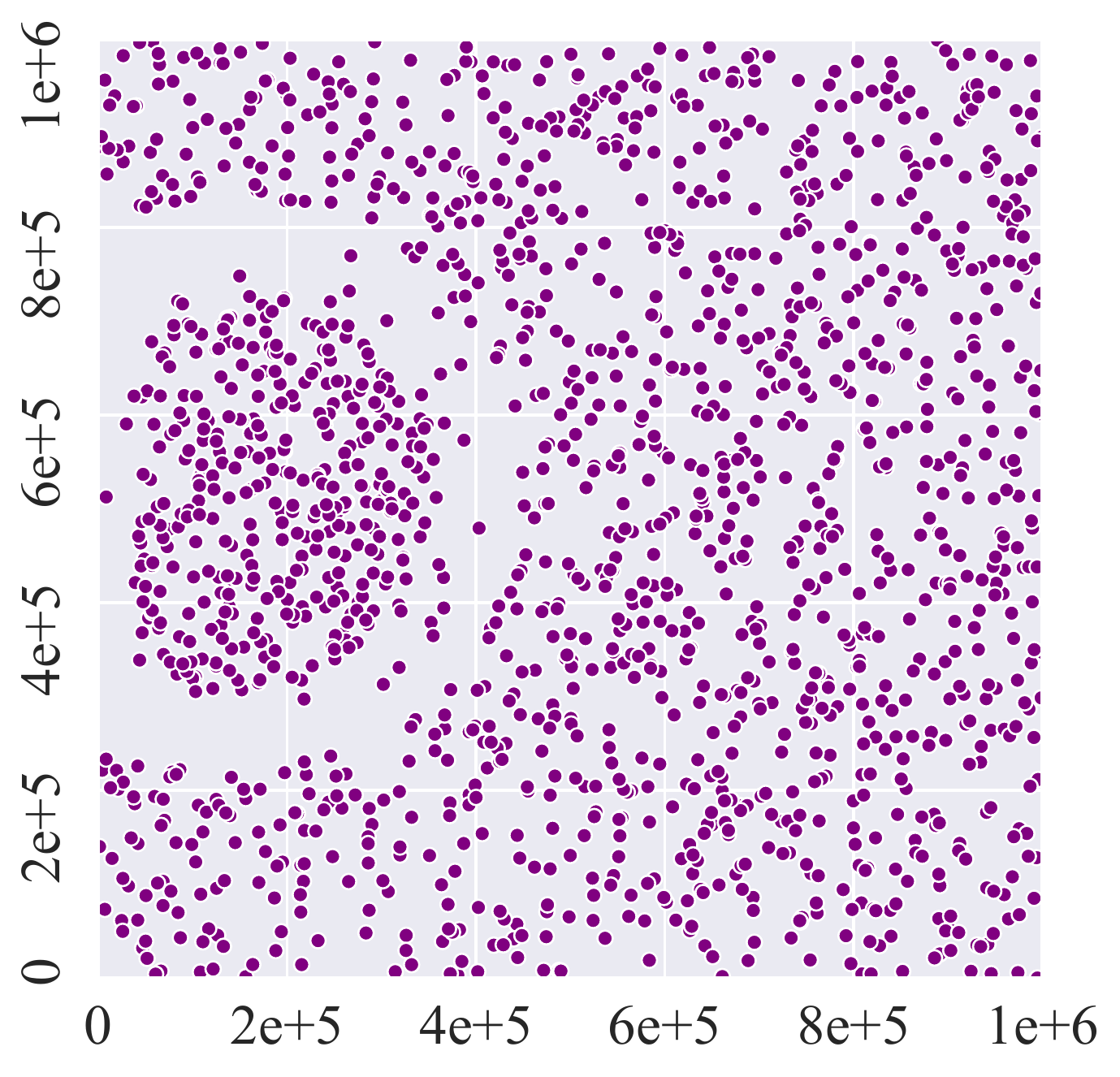}
   \caption{implosion}
   \label{fig:tsp:implosion}
\end{subfigure}
\hfill
\begin{subfigure}[b]{0.3\columnwidth}
  \includegraphics[width=\linewidth]{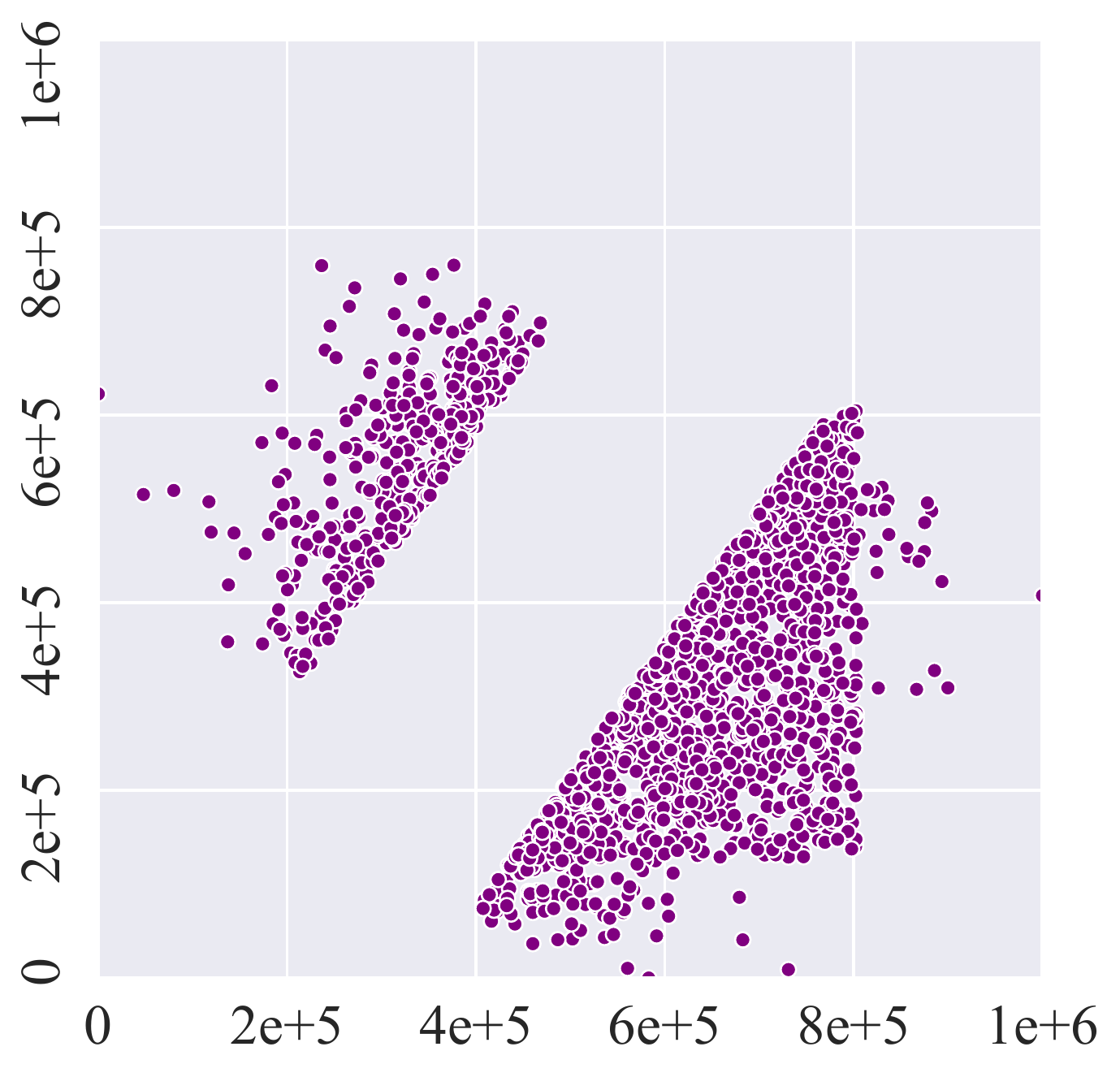} 
  \caption{expansion}
   \label{fig:tsp:expansion}
\end{subfigure}
\hfill
\begin{subfigure}[b]{0.3\columnwidth}
   \includegraphics[width=\linewidth]{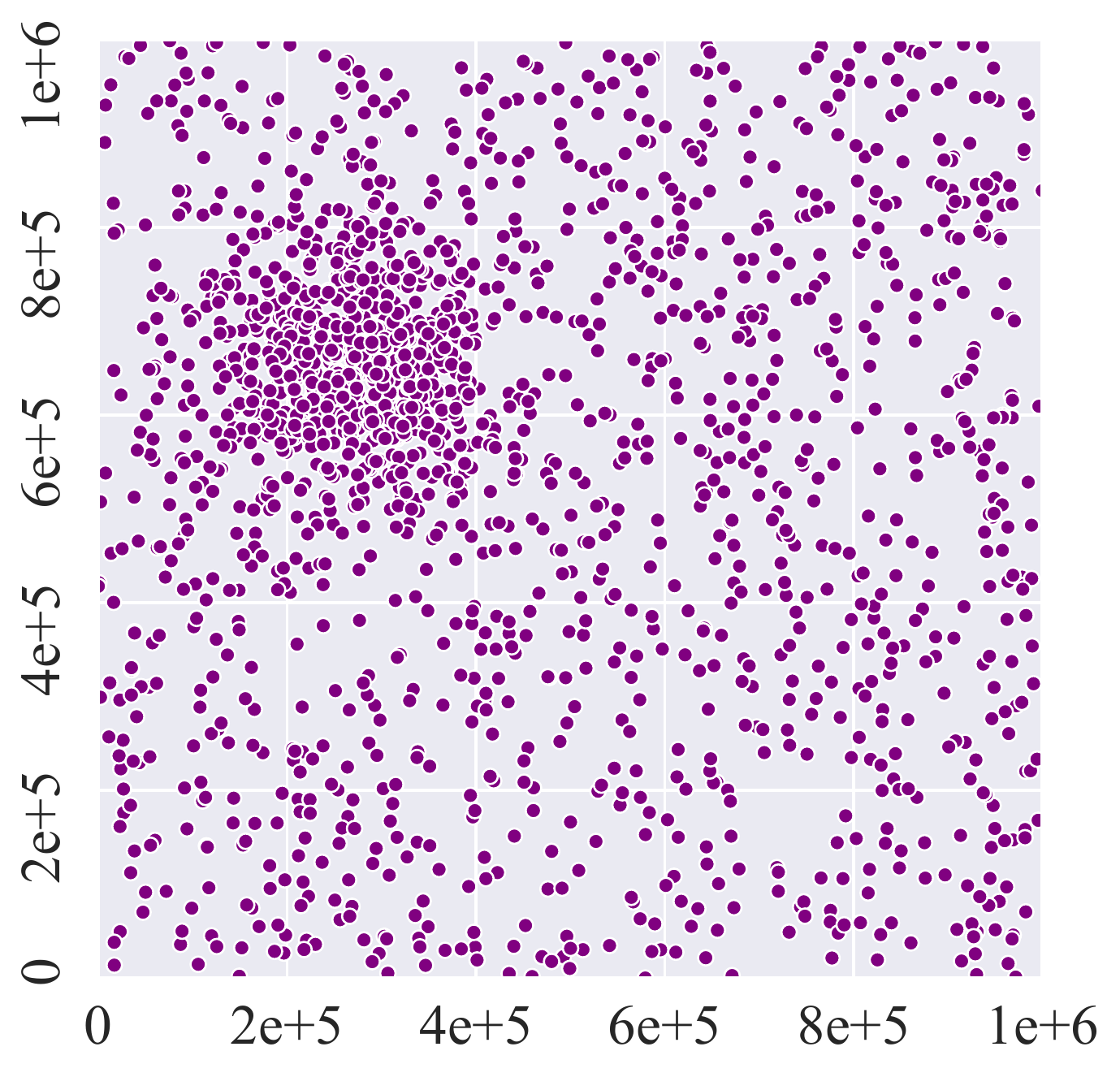}
   \caption{cluster}
   \label{fig:tsp:cluster}
\end{subfigure}
\hfill
\begin{subfigure}[b]{0.3\columnwidth}
  \includegraphics[width=\linewidth]{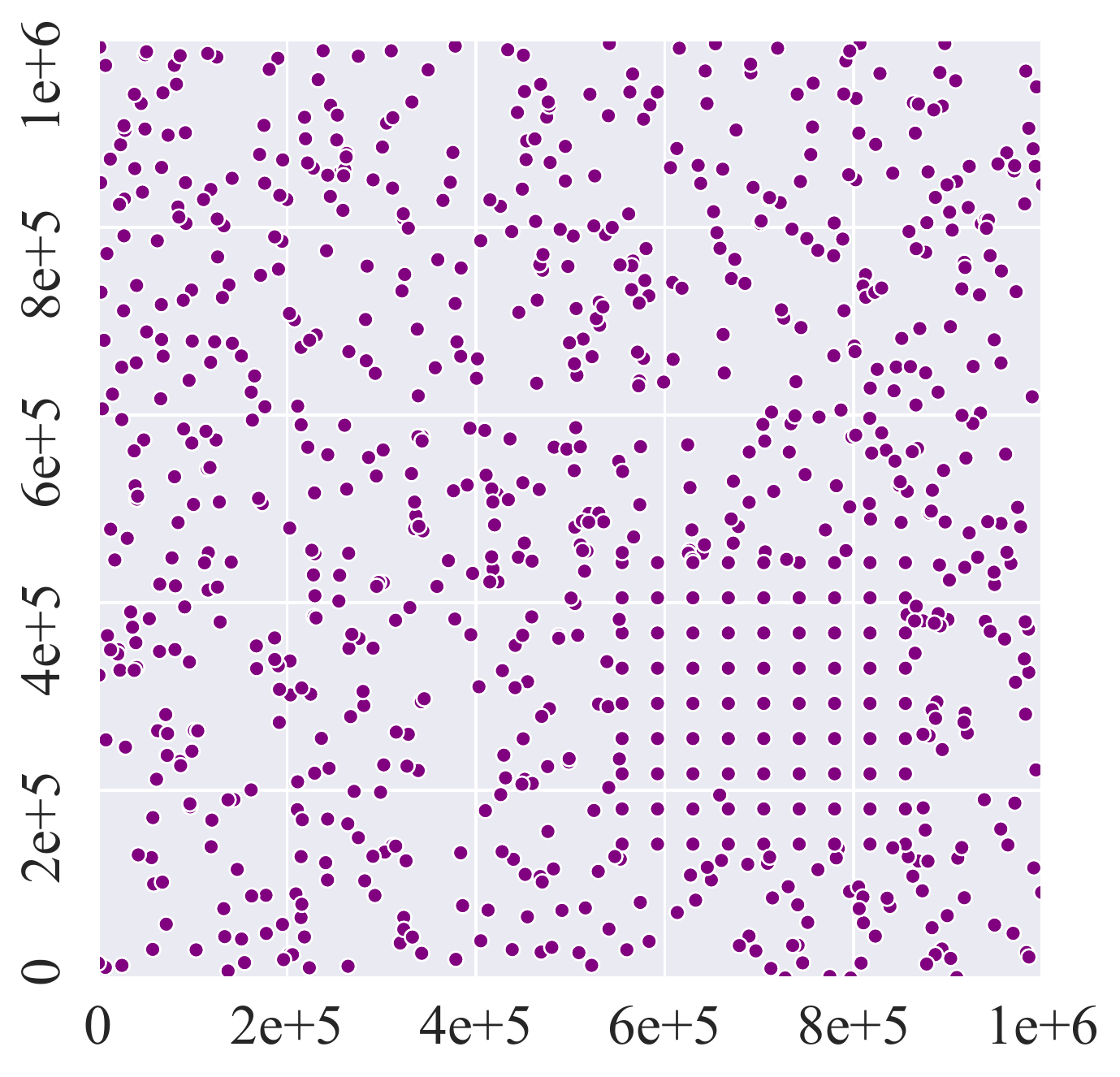}
  \caption{grid}
   \label{fig:tsp:grid}
\end{subfigure}
\caption{Visualization of the six types of TSP instances.}
\label{fig:tsp}
\end{figure*}

\subsection{Data Augmentation}
A large amount of training data is very important for obtaining a deep model with satisfactory generalization.
In this paper, we propose to enlarge the data diversity among the TSP density maps by data augmentation.
First of all, it is important to note that the safety of data augmentation, i.e., the likelihood of preserving the label post-transformation~\cite{ShortenK19}, must be assured.
Unlike conventional computer vision (CV) tasks, TSP is a combinatorial optimization problem and a solver to be identified is sensitive to local disturbance.
In other words, the data augmentation must strictly preserve the spatial distribution of the original points, in order to make the performance of TSP solvers consistent across the pre-transformed and post-transformed.
As a consequence, some commonly used augmentation strategies in CV, such as cropping, translation, and noise-injection, could not be used here.
Instead, we adopt two simple augmentation operators which keep the relative positions and the distances between all pairs of points unchanged.
The first one is flipping, which rolls an image over along a random straight line.
Fig.~\ref{fig:trans:hflip} and Fig.~\ref{fig:trans:vflip} show the results of flipping Fig.~\ref{fig:trans:interpolate} along $x = 0.5$ (horizontal flipping) and $y = 0.5$ (vertical flipping), respectively.
The second one is rotation, which rotates an image by an angle randomly sampled from $\{\frac{2\pi}{d}, 2\frac{2\pi}{d},\cdots,d\frac{2\pi}{d}\}$, where $d$ is a hyper-parameter.
Fig.~\ref{fig:trans:rotate} shows the result of rotating Fig.~\ref{fig:trans:interpolate} by $\frac{\pi}{2}$.
Both of the two operators are applied on the raw coordinates of a TSP instance, followed by the min-max normalization and gridding.
In CTAS, the CNN is trained by feeding randomly flipped and rotated density map.

\section{Data Collection}
\label{sec:benchmark}
The widely used TSP benchmark, TSPlib \cite{Reinelt91}, consists of around 100 instances, which is not enough for building a solver selector.
To support large-scale TSP solver selection, we construct a huge benchmark consisting of 6,000 instances using six different TSP generators, and collect six state-of-the-art TSP solvers.

\subsection{TSP Instances}
A diverse set of instances is important for exploiting the power of AS models. 
Here, diversity is relevant regarding three aspects \cite{BossekKN00T19}: the ability to discriminate performance of different TSP solvers, the topologies, as well as the coverage on the feature space.
To obtain such a set, we collect six different TSP generators, namely \textit{portgen}, \textit{explosion}, \textit{implosion}, \textit{expansion}, \textit{cluster} and \textit{grid}, from the literature.
Among them, \textit{portgen} generates a TSP instance (called rue instance) by uniform randomly placing the points on a Euclidean plane.
It has been used to create test beds for the 8th DIMACS Implementation Challenge \cite{johnson2007experimental}.
In comparison, the other five are proposed by a recent study \cite{BossekKN00T19} in which they have been shown effective to generate highly-diverse TSP instances.
As the names suggest, they generate a TSP instance mainly by simulating a phenomenon in the point clouds of a rue instance.
For example, the \textit{implosion} generator generates an instance by driving the points of a rue instance towards an implosion center. 
Fig.~\ref{fig:tsp} visualizes examples of the six types of instances, from which one could identify significant differences among them.
We generate 6,000 instances in total, specifically, 1,000 per type.
For each instance, the number of points is randomly sampled from [500,2000].

\begin{table}[t]
  \begin{center}
  \caption{Solver performance statistics over the instances.}
  \label{tbl:performance:analysis}
  \scalebox{0.75}{
     \begin{tabular}{l|lrrrrrr|r}
     \hline
     TSP set  & Measure &  EAX & EAXr & LKH & LKHr & LKHc & MAOS & VBS  \\
     \hline
     \hline
         \multirow{4}{*}{rue} & Unique  & 188 &153  & 204 & 127 & 223 & 81 & 1000  \\ 
                            & Shared  & 2 & 2 & 21 & 9 & 16  & 0 & 0 \\ 
                            & Failed  & 254 & 1 & 11 & 10 & 9 & 6 & 0 \\ 
                            & PAR10  & 2298.48 & 36.92 & 143.47 & 134.71 & 126.28 & 95.62 & 14.85 \\ \hline
     \multirow{4}{*}{explosion} & Unique  & 220 & 162 & 233 & 99 & 215 & 48 & 1000 \\ 
                            & Shared  & 4 & 4 & 17 & 3 & 18 & 0 & 0 \\ 
                            & Failed  & 194 & 1 & 5 & 3 & 3 & 5 & 0 \\ 
                            & PAR10  & 1758.07 & 30.14 & 84.72 & 66.89 & 65.73 & 77.50 & 12.72  \\ \hline
     \multirow{4}{*}{implosion} & Unique  & 215 & 152 & 238 & 106 & 199 & 49 & 1000 \\ 
                            & Shared  & 6 & 6 & 31 & 6 & 33 & 0 & 0 \\ 
                            & Failed  & 193 & 1 & 10 & 10 & 11 & 4 & 0 \\ 
                            & PAR10  & 1748.04 & 29.71 & 129.43 & 129.29  & 137.70 & 71.89 & 12.60 \\ \hline              
     \multirow{4}{*}{expansion} & Unique  &  299 & 214 & 10 & 9 & 8 & 451 & 1000 \\ 
                            & Shared  & 8 & 8 & 0 & 1 & 1 & 0 & 0 \\ 
                            & Failed  & 507 & 42 & 316 & 318 & 319 & 11  &  0 \\ 
                            & PAR10  & 4569.26  & 432.91 & 3001.87 & 3019.59 & 3026.38 & 123.02 & 19.67 \\ \hline                                         
     \multirow{4}{*}{cluster} & Unique  & 239 & 191 & 184 & 83 & 177 & 90 & 1000 \\ 
                            & Shared  & 6 & 6 & 24 & 9 & 27  & 0  & 0 \\ 
                            & Failed  & 246 & 1 & 54 & 55 & 53 & 7 & 0 \\ 
                            & PAR10  & 2225.21 & 34.35 & 546.62 & 555.76 & 541.13 & 100.63 & 14.51 \\ \hline
       \multirow{4}{*}{grid} & Unique  & 189 & 127 & 242 & 93 & 278 & 44 & 1000 \\ 
                            & Shared  & 4 & 4 & 20 & 5 & 21 & 0 & 0 \\ 
                            & Failed  & 234 & 1 & 15 & 13 & 14 & 34 & 0 \\ 
                            & PAR10  & 2117.78 & 43.78 & 177.67 & 160.81 & 168.64 & 364.00  & 15.28 \\ \hline                        
                                \hline
     \multirow{4}{*}{ Total} & Unique & 1350  & 999 & 1111 & 517 & 1100 & 763 & 6000  \\ 
                            & Shared  & 30 & 30  & 113 & 33 & 116 & 0 & 0 \\ 
                            & Failed  & 1628 & 47 & 411 & 409  & 409 & 67  & 0 \\ 
                            & PAR10  & 2452.81 & 101.30 & 680.63 & 677.84 & 677.64 & 138.78 & 14.94 \\ \hline 
\end{tabular}}
\end{center}
\end{table}

\subsection{TSP Solvers}
We consider the state-of-the-art approximate TSP solvers \cite{LiuTL020}, i.e., LKH, EAX and MAOS, and their variants.
Due to the rather different approaches underlying them, they are likely to achieve complementary strength in running time across extensive and diversified benchmarks.
LKH \cite{helsgaun2000effective} is a variant of the well-known Lin-Kernighan (LK) heuristic \cite{LinK73}, which generates local search moves by constructing a sequence of edge exchanges involving five or more edges.
Over decades, LKH has been widely recognized as the recommended method for finding high-quality solutions to a large variety of TSP instances. 
We use the latest version (2.0.9) of LKH.
Moreover, we also consider two variants of LKH, namely LKHr \cite{Dubois-LacosteH15} and LKHc, where the former is augmented with an additional restart mechanism, and the latter utilizes a genetic operator to generate initial solutions for local search process.
EAX \cite{nagata2013powerful} is a genetic algorithm equipped with a powerful edge assembly crossover.
After being improved several times, currently EAX can approach and even surpass the performance of LKH in solving a broad range of TSP instances.
Once again, we consider a variant of EAX augmented with a restart mechanism, namely EAXr, which triggers a restart whenever its original termination criterion is met.
The last considered solver, MAOS \cite{XieL09}, is a recently proposed multi-agent based TSP solver which does not contain any explicit local search heuristic.
All the six solvers considered here, i.e., LKH, LKHr, LKHc, EAX, EAXr, and MAOS, are modified to terminate when reaching a given solution quality or exhausting a given time budget.
Note that the design of CTAS is independent to the solver set. Thereby, it is flexible to plug new solvers into CTAS.

\subsection{Data Generation and Analysis}
\begin{figure}[t]
\centering
  \begin{subfigure}[b]{0.48\columnwidth}
   \centering
   \scalebox{1.0}{\includegraphics[width=\linewidth]{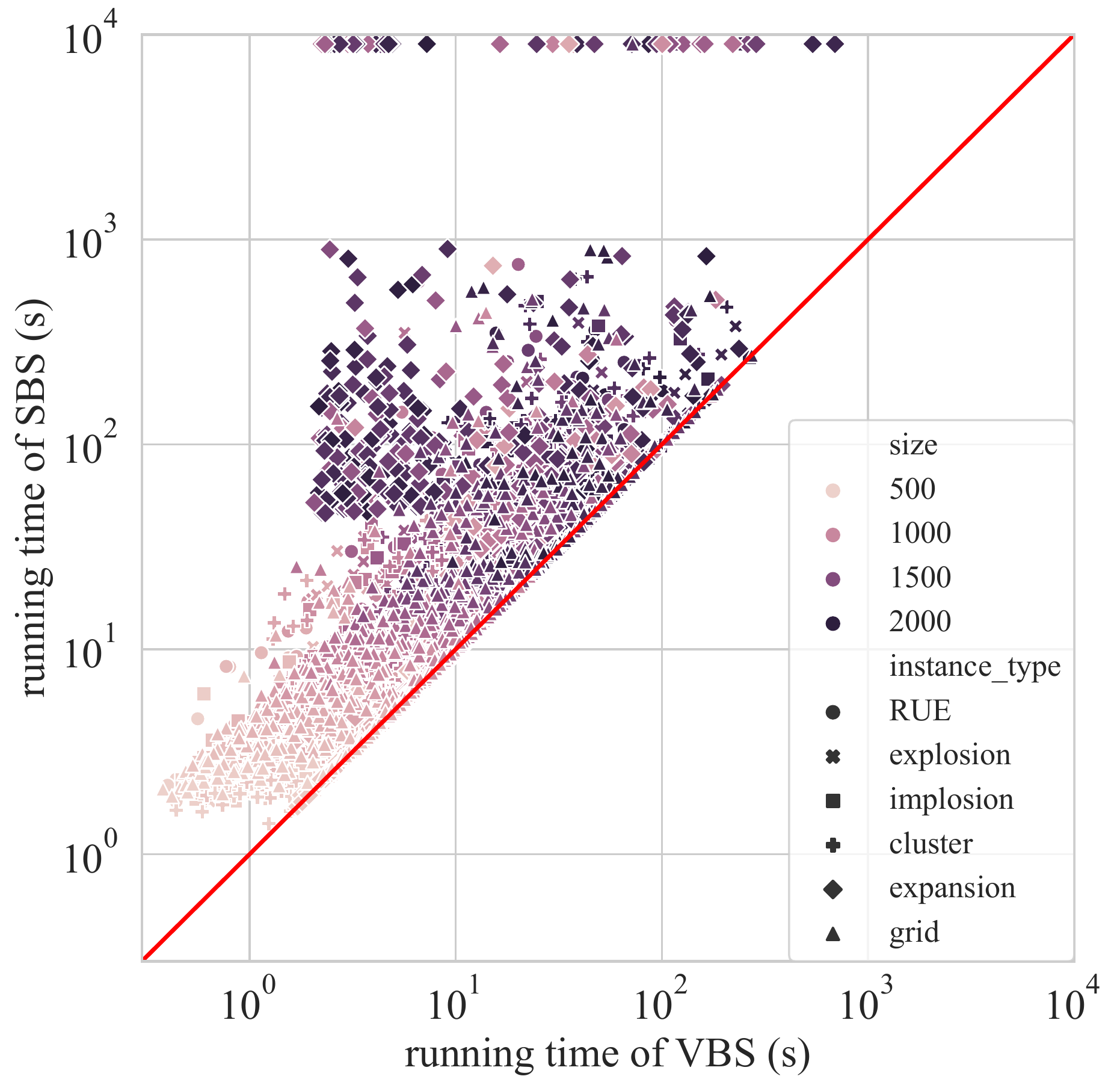}}
   \caption{VBS vs. SBS}
   \label{fig:solvers:vbs-sbs}
  \end{subfigure}
  \hfill
  \begin{subfigure}[b]{0.48\columnwidth}
   \centering
    \scalebox{1.0}{\includegraphics[width=\linewidth]{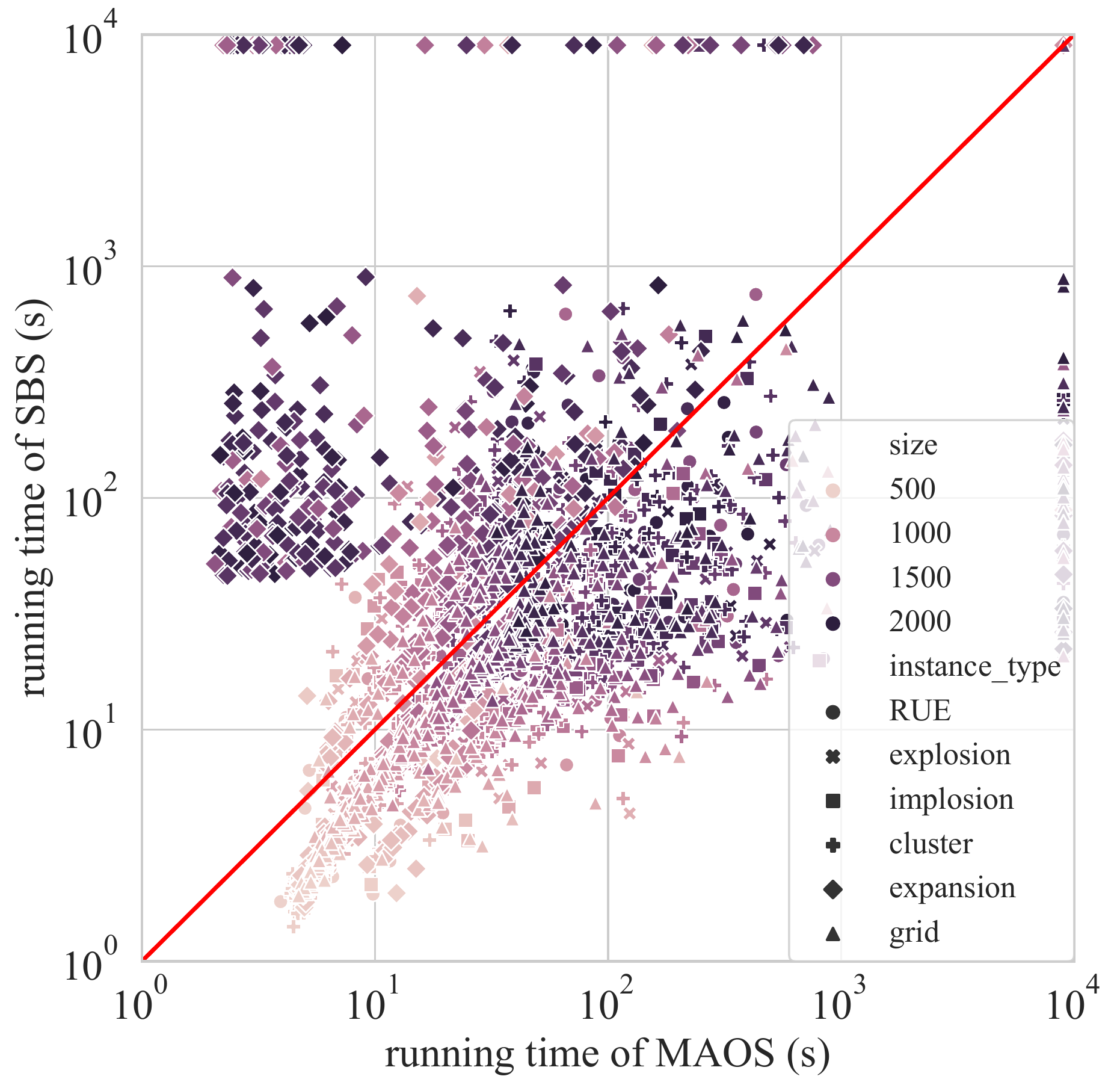}}
    \caption{SBS vs. 2nd best solver}
    \label{fig:solvers:maos-sbs}
  \end{subfigure}
\caption{Scatter plots of the running times.}
\label{fig:solvers}
\end{figure}

Since all the six solvers are randomized, we assess their performance based on five independent runs on each of the 6,000 instances.
Each run is terminated as soon as the solver finding the optimal solution (precomputed by Concorde \cite{applegate2006concorde}) 
or running for a cutoff time of 900 seconds;
in the first case, the run is considered successful and otherwise, unsuccessful.
The running time of each unsuccessful run is recorded as $10 \times 900s = 9000s$. 
After repeating five independent runs, the median running time over these runs is treated as the running time on the instance. 
The performance of a solver on an instance set is the average of its running time over all instances in the set, i.e., PAR10.
The performance statistics of the six solvers over the six types of TSP instances are shown in Table~\ref{tbl:performance:analysis}.
For each instance type, we distinguish the number of instances on which one solver is the unique one that performs best and multiple solvers achieve the same best performance, indicated as ``Unique'' and ``Shared'' in Table \ref{tbl:performance:analysis}, respectively.
In addition, the number of instances on which the solver fails to find the optimal solution is denoted as ``Failed''.

In general, LKH and its variant LKHr have the top number of unique best on grid, rue, explosion and implosion and share the best performance on a small fraction of instances.
The solver EAX achieves the most unique best on all the 6,000 instances.
Unfortunately, it also has the largest number of failed instances, resulting in the worst PAR10. 
For EAXr, although its performance is not the universal best on all the instance types, thanks to the limited number of failed instances, it achieves the minimum PAR10.
Thereby, EAXr will serve as SBS in the following study.
It is interesting to find that the restart mechanism plays an important role in boosting EAX, whereas it does not benefit much for LKH.
Only a small fraction of instances, 763 in 6,000, are best solved by MAOS, and more than half of these instances are from expansion.
However, the balanced performance of MAOS makes it become the 2nd best solver.
Fig.~\ref{fig:solvers} presents a fine-grind comparison of VBS with SBS (EAXr), and SBS with the 2nd best solver (MAOS).
As Fig.~\ref{fig:solvers:vbs-sbs} shows, there is a substantial performance gap between SBS and VBS across all instance types and sizes, which is exactly AS could deal with. 
Fig.~\ref{fig:solvers:maos-sbs} shows that among the 6,000 instances, MAOS outperforms SBS on 1145 instances with 464 from expansion.
Also, there are a small collection of instances in the other 5 types of instances that MAOS performs better. 

\section{Experimental Study}
\label{sec:exp}
Following the common scheme, in the experiments, we train the solver selectors based on a training set, and then evaluate them on an unseen test set.
We split the 6,000 instances 7:3 into training and test sets, by stratified sampling based on the partitions of the instance's best solver.
Below we describe the experimental settings. 

\subsection{Experimental Setup}
\subsubsection{CTAS Settings}
We used a state-of-the-art deep CNN, the residual networks (ResNet)~\cite{HeZRS16}, including the ResNet18 and ResNet34.
Equipped with the basic 3 $\times$ 3 filters as a shortcut connection, ResNet18 and ResNet34 have 18 and 34 layers, respectively.
The cross entropy loss (Eq.~(\ref{eq:loss:ce})) and the MSE (Eq.~(\ref{eq:loss:mse})) were used as loss functions for classification (Cla.) and regression (Reg.), respectively.
The weight $w_{ij}$ in Eq.~(\ref{eq:loss:ce}) and Eq.~(\ref{eq:loss:mse}) was set to $t_{ij}^\alpha$, where $\alpha$ is set to 0.5.
In the training phase, the same hyper-parameter setting was used for both Cla. and Reg., as presented in Table \ref{tbl:hyperpara}.
In the testing phase, the data augmentation, i.e., rotation and flipping, was disabled.  
We implemented CTAS based on PyTorch 1.3.1~\footnote{https://pytorch.org/} with Python 3.7, using Adam~\cite{KingmaB14} optimizer with a learning rate decay to train the CNN, and ran the experiments on an Nvidia V100.

\begin{table}[t]
\centering
\caption{The hyper-parameter settings.}
\label{tbl:hyperpara}
\scalebox{0.83}{
    \begin{tabular}{c|l|c}
    \hline
    \multicolumn{2}{c}{\bf Hyper-parameters}  & {\bf Values} \\
    \hline
    \hline
    \multirow{5}{*}{ResNet18$|$ResNet34} & learning rate & $10^{-4}|10^{-3}$\\ 
                   & mini-batch size   & $64$  \\ 
                  & \# epochs   & $50$ \\
                  & lr decay rate & $0.9|0.5$\\ 
                  & lr decay patience & 10 \\
                  \hline
   \multirow{4}{*}{\makecell{Data \\ Transformation}} &  \# grid $c$  & 64 \\
                      & Interpolation factor  & 4 \\
                                   & Random Flip  & True \\
                                   & \# Rotation directions ($d$) & $7|17$ \\ \hline
    \multirow{6}{*}{MLP} & \# hidden units & $128$ \\ 
                            & learning rate & $2^{-4}$\\ 
                   & mini-batch size   & $32$  \\ 
                  & \# epochs   & $50$  \\
                  & lr decay rate & $0.9$ \\ 
                  & lr decay patience & $10$ \\
                  \hline
\end{tabular}}
\end{table}

\subsubsection{Statistical Models Settings}
We compared our CNN with the state-of-the-art statistical models for TSP solver selection.
Specifically, we repeated the same settings in \cite{KerschkeKBHT18}, i.e., three learning strategies, three statistical models, four feature sets, and two feature selection approaches, resulting in 72 selectors trained in total.
More specifically, in addition to Cla. and Reg., a pairwise regression (P-Reg.) strategy was also tried.
This strategy trains a model to predict the performance difference between each pair of solvers (resulting in 15 models in total), and then selects the solver with the best predicted performance difference to all the other solvers on the given instance.
For each of the three learning strategy, decision trees (DT), random forests (RF), and support vector machines (SVM) were trained.
Each of these models was built using the Python package scikit-learn 0.23.2~\footnote{https://scikit-learn.org/} with its default settings.
Table~\ref{tbl:features} lists the statistics of the three feature sets, which were fed into the statistical models.
The union set of UBC and Pihera was also used, with 336 features in total.
Two feature selection approaches, namely sequential floating forward search (sffs) and sequential floating backward search (sbfs), were conducted on each of the 36 solver selectors, i.e., 3 learning strategies $\times$ 3 models $\times$ 4 feature sets, as mentioned above.
These two approaches were implemented based on Python package MLxtend 0.17.3~\footnote{http://rasbt.github.io/mlxtend/}.
During feature selection, the performance of the selectors was assessed by 5-fold cross-validation on the training set, based on PAR10 plus the feature computation time.
For more details on these selectors, interested readers may refer to the original paper \cite{KerschkeKBHT18}.
It is worth mentioning that the overall training and feature engineering for these selectors took one week on an Intel Xeon E5-2699A v4 machine with 44 cores and 256GB RAM. 

\begin{table}[t]
\centering
\small
\caption{Statistics of the feature sets.}
\label{tbl:features}
\scalebox{0.77}{
\begin{tabular}{l|r|r r r} \hline
    \multirow{2}{*}{\bf Feature Set}  & \multirow{2}{*}{\bf \# Features} & \multicolumn{3}{c}{\bf Computing Time (s)}  \\   
     &  &   {\bf Median} & {\bf Mean} & {\bf Stan. Dev.} \\  \hline \hline
   UBC~\cite{hutter2014algorithm}  & 49 & 9.00  & 11.74 & 8.66 \\ 
   UBC-cheap~\cite{hutter2014algorithm} & 12 & 0.12 & 0.14 & 1.00 \\ 
   Pihera~\cite{PiheraM14}  & 287 & 0.07 & 0.073 & 0.04 \\\hline
    \end{tabular}}
\end{table}

\subsubsection{Other Baselines}
We trained a two-layer Multilayer Perceptron (MLP) as the neural network baseline.
The input of MLP is the 336-dimensional union feature set of UBC and Pihera.
Similar to CTAS, it also used the cross entropy loss (Eq.~(\ref{eq:loss:ce})) for Cla. and the MSE (Eq.~(\ref{eq:loss:mse})) for Reg., and was trained by Adam optimizer.
Table \ref{tbl:hyperpara} lists the hyper-parameter settings of MLP.
In addition, we compared with the recently proposed deep learning TSP solver selector~\cite{SeilerPBKT20}, an 8-layer CNN model with Cla. strategy. Following the setting in~\cite{SeilerPBKT20}, original TSP instances are transformed to $512\times 512$ grid-scale images, and the hyperparameters are set as suggested in \cite{SeilerPBKT20}.

\begin{table}[t]
  \small
  \centering
  \caption{The test performance of CTAS and the baselines.}
  \label{tab:baseline}
    \scalebox{0.795}{
    \begin{tabular}{l l l|r r r r}
    \hline 
    \multicolumn{3}{c}{\bf Selector Characteristics} &     \multicolumn{4}{c}{\bf Performance}   \\
    \makecell{\bf Learn.\\ \bf Stra.} & \bf Model & \makecell{\bf Feat. Set / \\ \bf \#Used Feat.} & \makecell{\bf PAR10 \\ (s)} &  \makecell{\bf Avg. \\ \bf Rank} & \makecell{\bf Impro. \\ (\%)} & \makecell{\bf  Notwo.\\ (\%)}\\
    \hline \hline
    Reg. & DT    & Pihera   /4    & 41.01 & 3.08 & 7.50 & 91.17 \\
    Reg. & RF    & UBC-cheap    /8    & 58.39 & 3.05 & 14.89 & 85.22\\
    Reg. & SVM    & Pihera  /3    & 68.47 & 3.00 & 16.72 & 85.28 \\
    Cla. & DT & Pihera  /2  & 145.66 & 2.95 &  51.33 & 51.33 \\
    Cla. & SVM & UBC  /11  & 164.64 & 2.90 & 53.22 & 53.22\\
    Cla. & SVM & UBC-cheap /6 & 392.47 & 3.02 & 50.67 & 50.67\\
    P-Reg.  & SVM & UBC-cheap  /5 & 118.35 & 3.00 & 49.56 & 49.56 \\
    P-Reg. & SVM & UBC  /17  & 135.38 & 2.96 & 50.61 & 50.61 \\
    P-Reg.  & DT & Pihera  /2  & 147.05 & 4.28 & 17.83 & 17.83\\
    \hline 
    Reg. & MLP & union /336 & 52.87 & 2.17 & 7.06 & 88.89 \\
    Cla. & MLP & union /336 & 57.66 & 2.12 & 7.22 & 90.67 \\ 
    Cla. & 8-layer CNN & - & 111.03 & 1.92 & 32.00 & 50.50 \\ 
    \hline 
    Reg. & ResNet18 & - & {\bf 38.23} & 2.30 &  7.70 & 80.17  \\
    Reg. & ResNet34 & - & {\bf 36.89} & 1.92 &  18.56 & 86.83 \\
    Cla. & ResNet18 & - & 40.91 & 2.11 & 6.70 & 92.09 \\
    Cla. & ResNet34 & - & 45.93 & 2.04 & 8.17 & 93.83 \\
    \hline 
    - & SBS & - & 97.02 &  2.09 & 0.00 & 100.00 \\
    - & VBS & - & 13.90 & 1.00 & 86.10 & 100.00 \\
    \hline
    \end{tabular}}
\end{table}

\subsection{Results and Analysis}

Table~\ref{tab:baseline} summarizes the overall performance of the statistical models, MLP and the ResNets of CTAS.
Due to the limited space, for each learning strategy, Table \ref{tab:baseline} only lists the top-3 statistical models regarding the test PAR10.
The feature sets of all these nine selectors are obtained by sffs.
Note that for a feature-based selector (i.e., DT, RF, SVM and MLP), its performance on a given TSP instance is the running time of the selected solver plus the running time required for computing the instance features used by the selector.
Compared to the solver running time and feature computation time, the model prediction time is negligible. 
Averaged on each instance, model prediction of the ResNets spend about $10^{-3}$ second on CPU and $10^{-4}$ second on GPU, and the statistical models as well as MLP need $10^{-3} \sim 10^{-7}$ second, depending on the model, learning strategy and the number of used features.
Meanwhile, we list the average rank of the selected solver (Avg. Rank), as well as the percentage of the test instances on which the selected solver is better (Impro.) and not worse (Notwo.) than SBS (EAXr).
Note that there is no strict correlation between PAR10 and Avg. Rank (neither Impro. nor Notwo.), since PAR10 involves the discrepancy of solvers' running time and is more sensitive to timeout cases while Avg. Rank, Impro. and Notwo. reflect
a rough selection preference of the selector.

\subsubsection{Different Models} 
In general, the Reg. ResNets achieve over 2$\times$ speedup in PAR10 comparing with SBS, and 
outperform the baselines in both PAR10 and Avg. Rank.
Such results indicate that deep CNN is able to extract useful features from the density maps of TSP instances.
The top baselines Reg. DT(Pihera) and Reg. RF(UBC-cheap) have two and three timeouts, respectively, while the Reg. ResNet18 and Reg. ResNet34 both have only 1 timeout.
On the other hand, the 8-layer CNN baseline performs significantly worse than the best statistical methods, due to its uninformative visual representations of TSP instances and inappropriate learning strategy (see the discussion below).
It can be seen that all classical AS models only use a small fraction of the available features.
In other words, a majority of the features are actually not useful for improving the performance of the solver selector.
In fact, in our experiments, we found none of the models using all the 336 features were able to beat SBS.
This reflects that, as aforementioned, feature engineering is essential for hand-crafted feature-based solver selection systems. 

\begin{figure*}[!ht]
\centering
\begin{subfigure}[b]{0.48\columnwidth}
  \centering
  \scalebox{0.93}{\includegraphics[width=\linewidth]{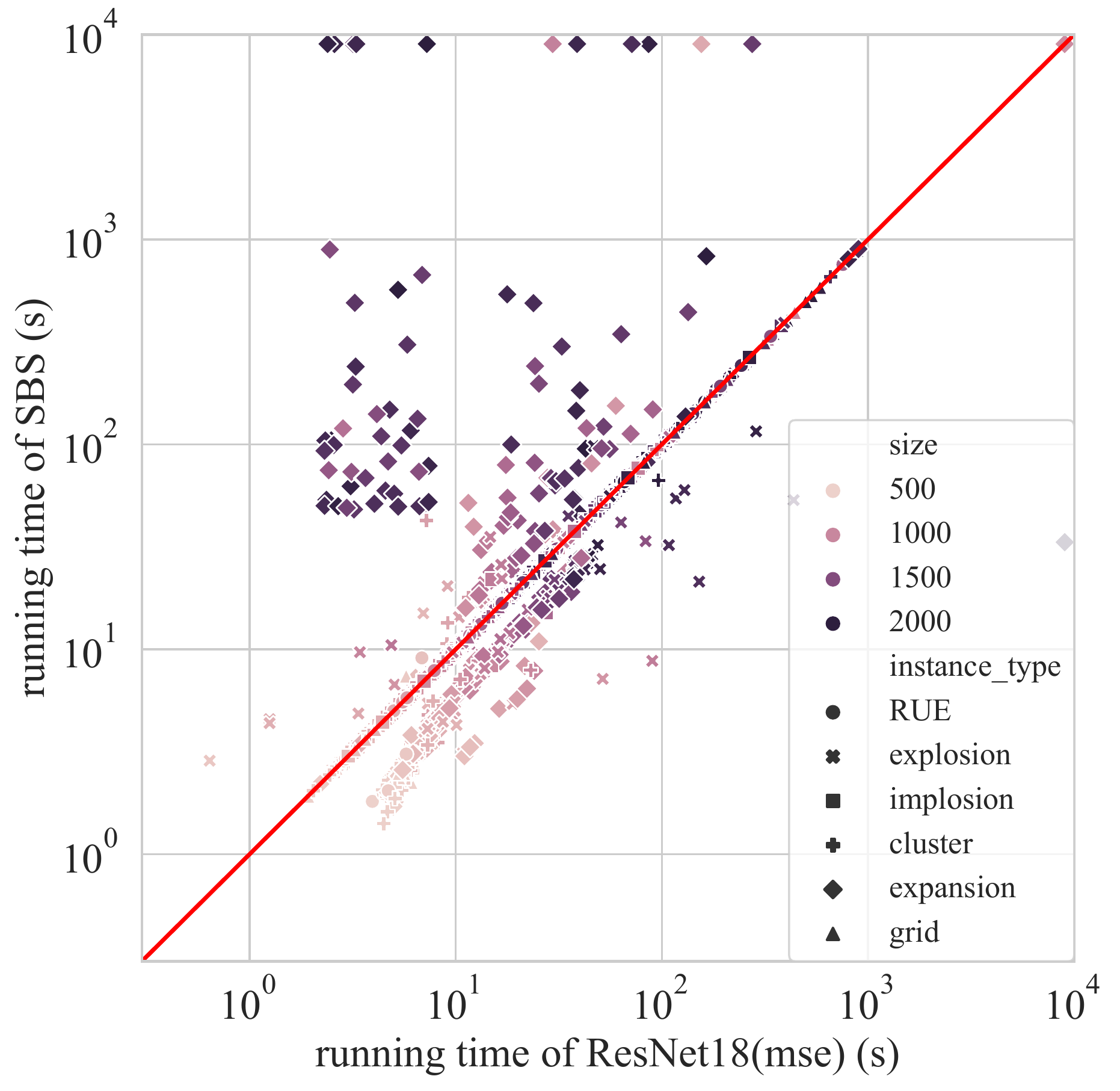}}
  \caption{SBS vs. Reg. ResNet18}
  \label{fig:res:resnet18mse-sbs}
\end{subfigure}
\hfill
\begin{subfigure}[b]{0.48\columnwidth}
  \centering
  \scalebox{0.93}{\includegraphics[width=\linewidth]{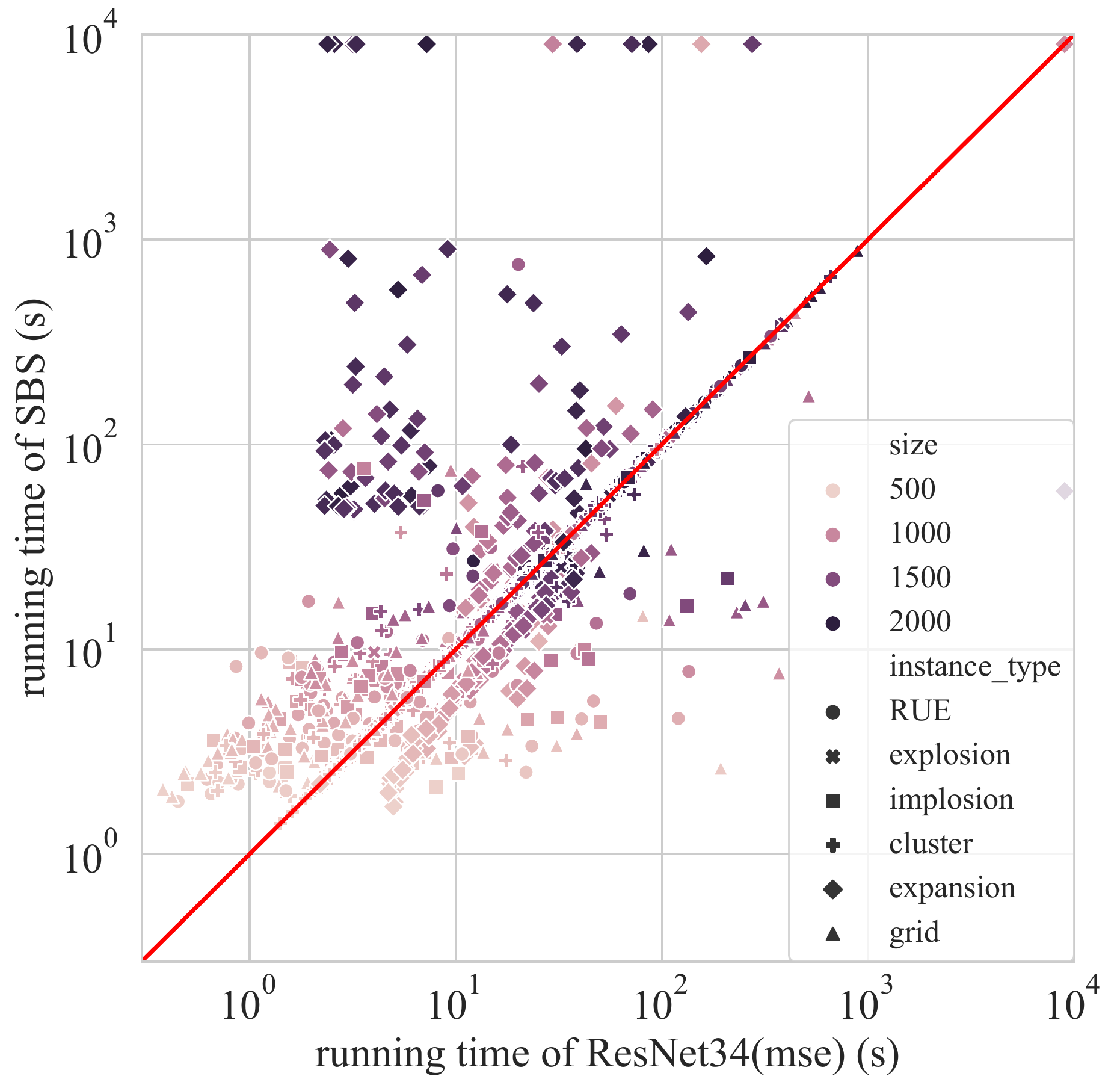}}
  \caption{SBS vs. Reg. ResNet34}
  \label{fig:res:resnet34mse-sbs}
\end{subfigure}
\hfill
\begin{subfigure}[b]{0.48\columnwidth}
  \centering
  \scalebox{0.93}{\includegraphics[width=\linewidth]{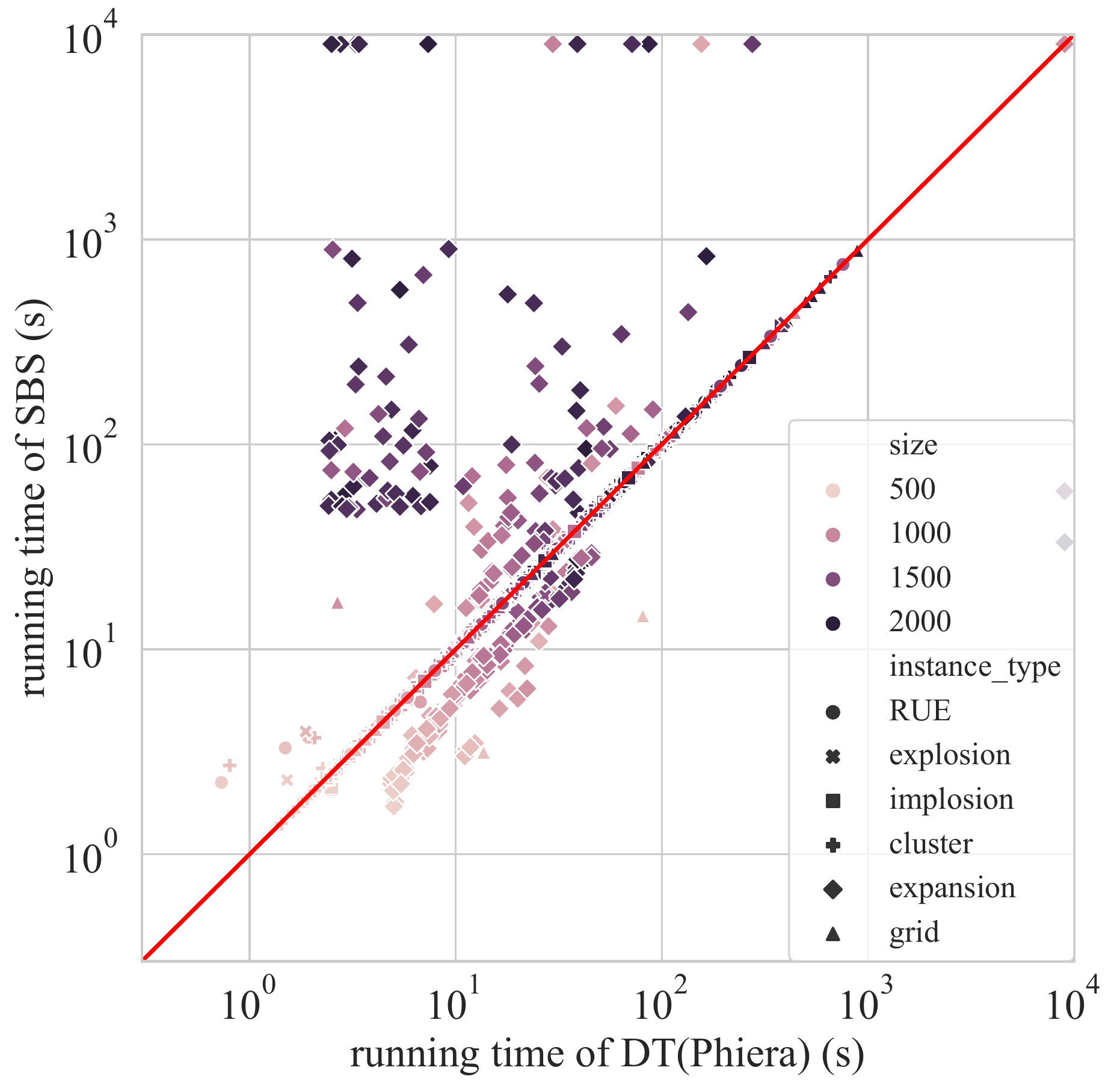}}
  \caption{SBS vs. Reg. DT}
  \label{fig:res:dt-phiera-sbs}
\end{subfigure}
\hfill
\begin{subfigure}[b]{0.48\columnwidth}
  \centering
  \scalebox{0.93}{\includegraphics[width=\linewidth]{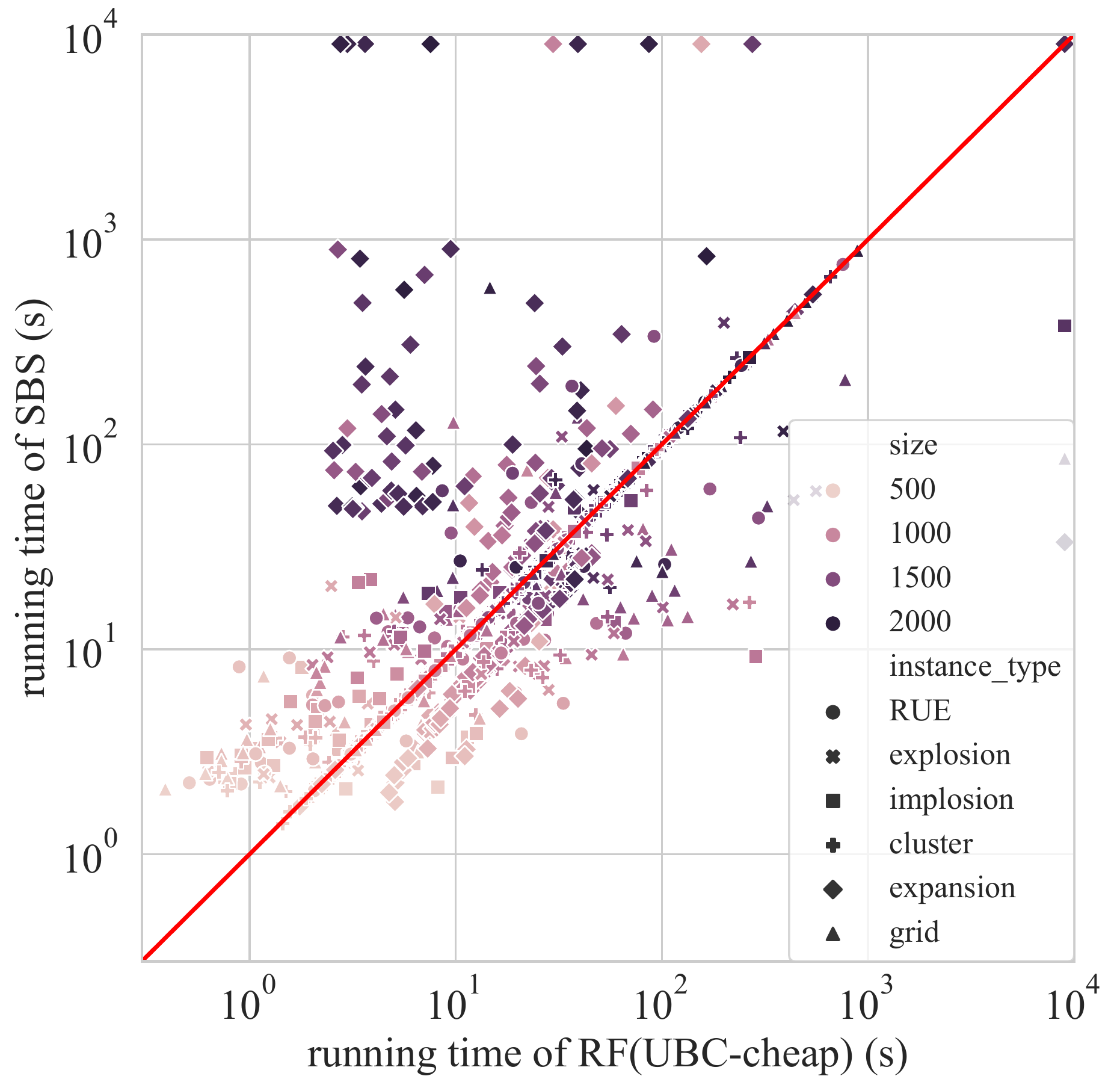}}
  \caption{SBS vs. Reg. RF}
  \label{fig:res:rf-ubccheap-sbs}
\end{subfigure}
\caption{Scatter plots of the running times over the 1800 test instances.}
\label{fig:allresults}
\end{figure*}

\subsubsection{Different Learning Strategies}
Another important observation from Table~\ref{tab:baseline} is that the regression (Reg.) strategy significantly outperforms the classification (Cla.) strategy, and the result is consistent across all the selectors. 
As discussed before, Cla. strategy solely relies on matching the empirical distribution of the best solvers so that the Cla. models tend to achieve relatively higher Avg. Rank and the improvement ratio.
However, the cross entropy loss does not take the magnitude of the discrepancy of running time into account. 
The instance-wise weights in Eq.~(\ref{eq:loss:ce}) can only alleviate rather than solve this problem. 
In contrast, Reg. strategy directly predicts the running time of the solvers, which leverages more information than classification, especially for the powerful deep learning models, e.g., ResNet18 and ResNet34.
The performance of pairwise regression (P-Reg.) selectors is between Cla. selectors and Reg. selectors.
As one P-reg. solver selector is composed of 15 models whose prediction is the difference of solvers' running time, we conjecture that this introduces more uncertainty and errors in the selector. 

\subsubsection{Selector Visualization}
To make a close observation, we visualize the selection results for the best-performing Reg. selectors in Fig.~\ref{fig:allresults}, compared with SBS.
The points falling exactly on the red slash line indicate the instances for which the corresponding selector predicts SBS.
The points above and below the line indicate that the selected solver performs better and worse than SBS, respectively. 
All the four selectors in Fig.~\ref{fig:allresults} attempt to achieve improvement over SBS on the large-scale instances (dark points in Fig.~\ref{fig:allresults}), with the sacrifice of performance on the small-scale instances.
This is reasonable for a solver selector to optimize PAR10 because making right decisions on large-scale instances has a larger effect on PAR10 than on small-scale ones, considering both the mean and variance of the solver running times will get larger as the number of points in instances increases.
For a solver selector, there is a subtle trade-off between keeping the choice of SBS and trying to recommend a better solver. 
The former never incurs performance degradation while the latter brings the chance of improvement at the risk of degradation.
The best statistical baseline Reg. DT(Pihera) (Fig.~\ref{fig:res:dt-phiera-sbs}) adopts the former conservative strategy.
It selected SBS for 83.67\% test instances ($91.17\% - 7.50\%$, see Table~\ref{tab:baseline}). 
In comparison, the corresponding ratio for Reg. ResNet18 and Reg. ResNet34 are 72.47\% and 68.27\%, respectively.
In fact, on many grid instances, Reg. ResNets tended to select solvers from the LKH solver family, which indeed performed better than SBS on these instances (see Fig.~\ref{fig:res:resnet18mse-sbs} and Fig.~\ref{fig:res:resnet34mse-sbs}).
On the other hand, a non-conservative strategy might also make overall performance deteriorate.
For example, Reg. RF(UBC-cheap) (Fig.~\ref{fig:res:rf-ubccheap-sbs}) selected SBS for 70.33\% test instances, smaller than Reg. ResNet18.
However, it also resulted in three timeouts, which substantially increased the overall PAR10.

\begin{figure}[t]
\centering
\begin{subfigure}[b]{0.48\columnwidth}
  \centering
  \scalebox{0.93}{\includegraphics[width=\linewidth]{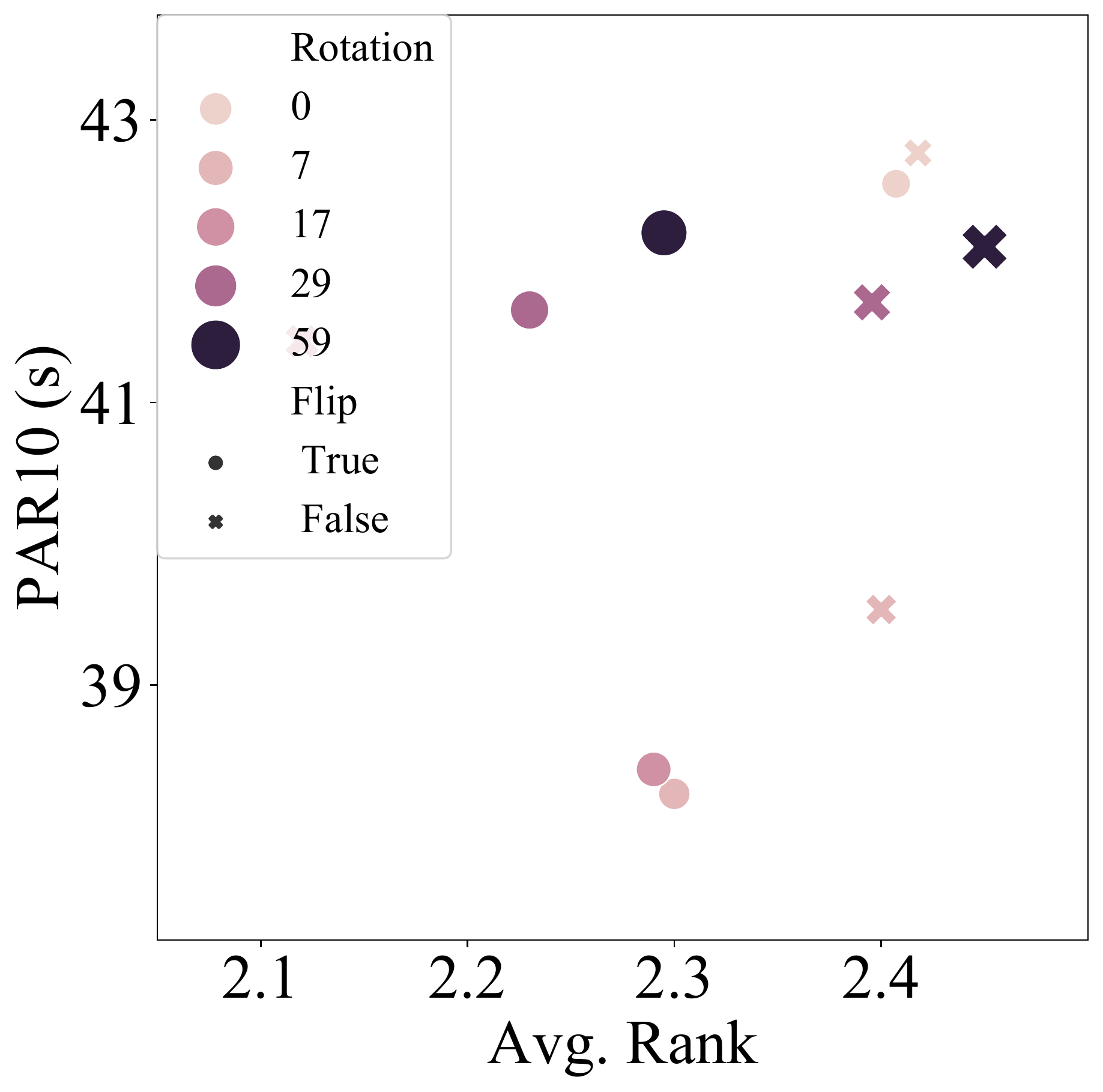}}
  \caption{Reg. ResNet18}
  \label{fig:aug:resnet18}
\end{subfigure}
\hfill
\begin{subfigure}[b]{0.48\columnwidth}
  \centering
  \scalebox{0.93}{\includegraphics[width=\linewidth]{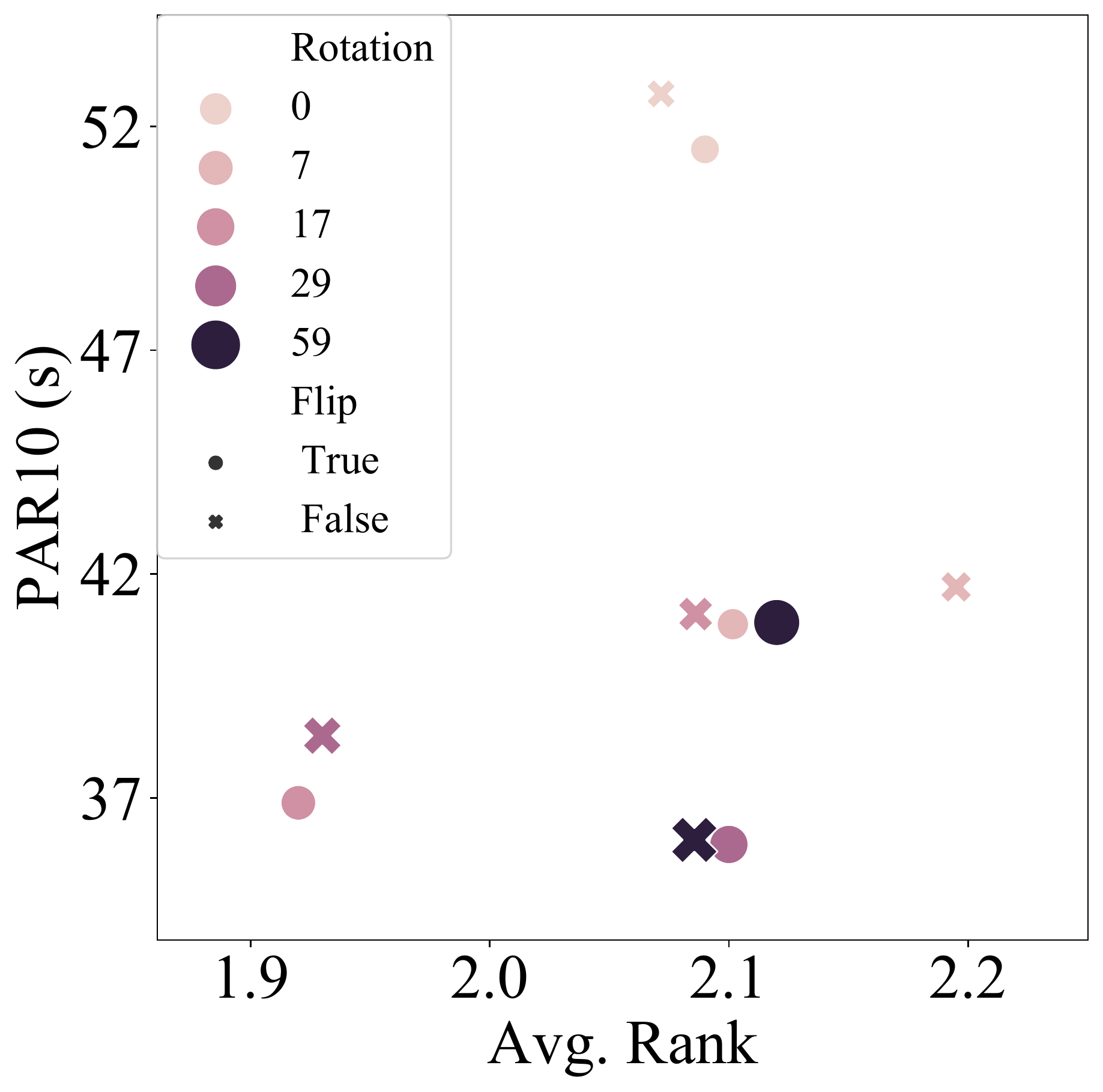}}
  \caption{Reg. ResNet34}
  \label{fig:aug:resnet34}
\end{subfigure}
\caption{The effect of flipping and rotation.}
\label{fig:exp:augmentation}
\end{figure}

\subsection{Effect of Data Augmentation}
We conduct an ablation study to validate the effectiveness of the data augmentation strategies.
Fig.~\ref{fig:aug:resnet18} and Fig.~\ref{fig:aug:resnet34} show the PAR10 and the Avg. Rank
of the Reg. ResNets, when varying the degree of random rotation and turning on/off flipping. 
The size and color of the points indicate the number of directions (i.e., $d$) associated with the rotation operator.
'$\bullet$' and '$\times$' indicate the flipping is enabled and disabled, respectively.
The models were trained with identical hyper-parameters as before, except the data augmentation.
The most important finding from Fig.~\ref{fig:exp:augmentation} is augmentation is essential for our task. 
At the top-right of Fig.~\ref{fig:exp:augmentation}, training without rotation ($d=0$) or flipping leads to large PAR10 and low Avg. Rank, for both ResNet18 and ResNet34.  
As we increase the number of rotation directions and turn on flipping, the test performance could achieve its best. 
Compared with ResNet18, ResNet34 trained with more rotation directions will obtain the best performance. That is in accordance with our intuition that the larger the model, the more diverse data it requires. 

\section{Conclusion}
\label{sec:conclusion}
In this paper, we propose an end-to-end learning framework, CTAS, for TSP solver selection.
We utilize deep CNN with data augmentation, and have explored different learning strategies.
Extensive experiments have shown the superiority of CTAS over SBS and classical AS models, which may spend days to obtain the best feature set.
The work thus shows the huge potential of deep learning in this domain.
Future directions include specially designed loss functions and network structures for AS tasks, and adapting CTAS to solver selection in vehicle routing problems \cite{liu2020memetic}.

\section*{Acknowledgement}
This work is supported by the Research Grants Council of Hong Kong, China under No. 14203618, No. 14202919 and No. 14205520.

\bibliography{mybib}

\end{document}